\documentclass[journal]{IEEEtran}
\usepackage{amsfonts,array}
\usepackage[dvipsnames,usenames]{color}
\usepackage{graphicx,times,psfig,amsmath} 
\usepackage{graphics,color}
\usepackage{amssymb,amsmath,amsfonts} 
\usepackage{amsmath,mathrsfs,bm,url,times}
\usepackage{latexsym}
\usepackage{subfigure,float}
\usepackage{algorithm}
\usepackage{algorithmicx}
\usepackage{algpseudocode}
\usepackage{epsfig,url}
\usepackage{epstopdf}
\usepackage{multirow}
\usepackage{stmaryrd}

\newtheorem{proposition}{Proposition}
\newtheorem{theorem}{Theorem}
\newtheorem{lemma}{Lemma}

\newtheorem{definition}{Definition}

\newtheorem{remark}{Remark}

\DeclareMathOperator*{\argmax}{argmax}
\newcommand{\R}{{\mathbb R}}
\newcommand{\onetom}{1,2,\cdots,m}
\newcommand{\oneton}{1,\cdots,n}

\newcommand{\zerotoinfty}{0,1,2,\cdots}

\newcommand{\measure}{\text{\it m}}
\newcommand{\ignore}[1]{}

\begin{document}
\title{Stability of Analytic Neural Networks with Event-triggered Synaptic Feedbacks}
\author{Ren Zheng, Xinlei Yi , Wenlian Lu~\IEEEmembership{Senior~Member,~IEEE}, Tianping Chen~\IEEEmembership{Senior~Member,~IEEE}\thanks{This manuscript is submitted to the Special Issue on Neurodynamic Systems for Optimization and Applications.}\thanks{R. Zheng, X. Yi, W. L. Lu and T. P. Chen are with the School of Mathematical Sciences, Fudan University, China; W. L. Lu is also with the Centre for Computational Systems Biology, Fudan University; T. P. Chen is also with the School of Computer Science, Fudan University, China (email: \{yix11, wenlian, tchen\}@fudan.edu.cn).} \thanks{This work is jointly supported by the National Natural Sciences Foundation
of China (Nos. 61273211 and 61273309), the Program for New Century Excellent Talents in University (NCET-13-0139), the Programme of Introducing Talents of Discipline to Universities (B08018), and the Laboratory of Mathematics for Nonlinear Sciences
School of Mathematical Sciences, Fudan University.}}


\date{}

\maketitle

\begin{abstract}
In this paper, we investigate stability of a class of analytic neural
networks with the synaptic feedback via event-triggered rules. This model is general and include Hopfield neural network as a special case. These event-trigger rules can efficiently reduces loads of computation and information transmission at synapses of the neurons. The synaptic feedback of each neuron keeps a constant value based on the
outputs of the other neurons at its latest triggering time but changes at its next
triggering time, which is determined by certain criterion. It is proved that every trajectory of the analytic neural
network converges to certain equilibrium under this event-triggered rule for all initial values except a set of zero measure. The main
technique of the proof is the {\L}ojasiewicz inequality to prove the finiteness of trajectory length. The realization of this event-triggered rule
is verified by the exclusion of Zeno behaviors. Numerical examples are
provided to illustrate the efficiency of the theoretical results.
\end{abstract}

\section{Introduction}
\PARstart{R}{ecurrently} connected neural networks have been extensively studied, and many applications
in different areas have been arising. Such applications heavily
depend on the stable dynamical behavior of the networks. For example, in application for optimisation, convergence of dynamics is fundamental, which has attracting many interests from different fields. See \cite{Mac}-\cite{Tpcl} and the references therein. Therefore, analysis of
these behaviors is a necessary step for practical design of neural networks.

This paper focuses on the following dynamical system
\begin{align}
\label{mg0.1}
\begin{cases}
\dot{x}=-D x-\nabla f(y)+\theta\\[3pt]
y=g(\Lambda x),\\[2pt]
\end{cases}
\end{align}
where $x\in\R^n$ is the state vector, $D=diag\{d_{1},\cdots,d_{n}\}$ with $d_{i}>0$ for all $i=\oneton$ is the self-inhibition matrix, the cost function $f(y):\R^{n}\to\R$ is an {\em analytic} function and $\theta\in\R^n$ is a constant input vector. $y=g(\Lambda x)$ is the output vector with the sigmoid function $g(\cdot)$ as nonlinear activation function and the scaling slopes $\Lambda=diag\{\lambda_{1},\cdots,\lambda_{n}\}$ for some $\lambda_{i}>0$ for all $i=\oneton$.

Eq. \eqref{mg0.1} was firstly proposed in \cite{Mfa} and is a general model of neural network system arising in recent years. For example, the well-known Hopfield neural network \cite{Jjh1,Jjh}, whose continuous-time version can be formulated as
\begin{align}
\begin{cases}
C_i\dot{x}_{i}=-\dfrac{x_{i}}{R_{i}}+\sum\limits_{j=1}^{n}\omega_{ij}y_{j}+\theta_{i}\\[3pt]
y_{i}=g_{i}(\lambda_{i}x_{i}),\\[2pt]
\end{cases}\label{mg0.2}
\end{align}
for $i=\oneton$, where $x_{i}$ stands for the state of neuron $i$ and each activation function $g_{i}(\cdot)$ is  sigmoid. With the symmetric weight condition ($\omega_{ij}=\omega_{ji}$ for all $i,j=1,\cdots,n$), Eq. (\ref{mg0.2}) can be formulated as Eq. (\ref{mg0.1}) with $f(y)=-\frac{1}{2}\sum_{i,j=1}^{m}\omega_{ij}y_iy_j$. This model has a great variety of applications. It can be used to search for local minima of the quadratic objective function of $f(y)$ over the discrete set $\{0,1\}^{n}$ \cite{Mvi}-\cite{Wll}, for example, the traveling-sales problem \cite{Jma}.
This model can be regarded as a special form of (\ref{mg0.1}) with $f(y)=E(y)$ and proved to minimize $E(y)$ over the discrete set $\{0,1\}^{n}$ \cite{Mvi}.

The linearization technique and the
classical LaSalle approach for proving stability (See \cite{Mac,Mvi}) could be invalid when the system had non-isolated equilibrium
points (e.g., a manifold of equilibria) \cite{Mfa}. The concept
"absolute stability" was proposed in \cite{Mfa1}-\cite{Mfa2} to show that
each trajectory of the neural network converges to certain equilibrium for any parameters and initial values by proving the finiteness of the
trajectory length and the celebrated {\L}ojasiewicz inequality
\cite{Slo}-\cite{Slo1}. This idea was also seen in \cite{Tpc}.

However, in the model (\ref{mg0.1}), the synaptic feedback of each neuron is simultaneous according to the output states of the other neurons, which is costly in practice for a network of a large number of
neurons. In recent years, with the development of sensing, communications,
and computing equipment, event-triggered control
\cite{Pta}-\cite{Yfg} and self-triggered control \cite{Aapt}-\cite{Wzh2}
have been proposed and have remarkable advantages that reduce the frequency of synaptic information exchange significantly.
In this paper, we investigate stability of analytic
neural networks with event-triggered synaptic feedbacks. Here, we present an
event-triggered rule to reduce the frequency of receiving synaptic
feedbacks. At each neuron, the synaptic feedback is a constant determined by the outputs of
the neurons at its latest triggering time and changes at the next triggering time
of this neuron that is triggered by a criterion via the neurons' output states
as well. We prove that the analytic neural network system is
{\em almost sure stable} (see {\it Definition \ref{convergence}}), which was proposed by Hirsh \cite{Mhi}, under the event-triggered rule by the {\L}ojasiewicz inequality. In addition, we further prove that the event-triggered rule is physically viable, owing to the
exclusion of Zeno behaviors. For the event-triggered rule,
each neuron needs the states of the other neurons and itself,
asynchronous. Hence, the neurons are not triggered in a
synchronous way, but independent of each other. It should be
highlighted that our results can be extended to a large class of
neural networks, for example, the standard cellular networks
\cite{Loc}-\cite{Loc1}.

The paper is organized as follows. In Section \ref{sec2}, the preliminaries are given; Stability and the exclusion of Zeno behaviours of analytic neural networks with the event-triggering rules are proved in Section \ref{sec3}; Then the discrete-time monitoring scheme is discussed in Section \ref{Monitoring}; In Section \ref{sec5}, numerical examples are provided to show the effectiveness of the
theoretical results; The paper is concluded in Section \ref{sec6}.

\noindent {\bf Notions}: $\R^{n}$ denotes $n$-dimensional real space. $\|\cdot\|$ represents the Euclidean norm for vectors or the induced 2-norm for matrices. $B_r(x_0)=\{x\in\R^{n}:~\|x-x_0\|<r\}$ stands for an $n$-dimensional ball
with center $x_0\in\R^{n}$ and radius $r>0$. For a function
$F(x):~\R^{n}\rightarrow\R$, $\nabla F(x)$ stands for its gradient. For a set $Q\subseteq \R^{n}$ and a point $x_0\in\R^{n}$,
$\text{\it dist\,}(x_0,Q)=\inf_{y\in Q}\|x_0-y\|$ indicates the distance from $x_0$ to
$Q$. $m(\cdot)$ stands for the Lebesgue measure in $\R^{n}$.

\section{Preliminaries and problem formulation}\label{sec2}

In this section, we firstly provide some definitions and preliminary results, which will be used later. With the discontinuous synaptic feedback, we consider Eq. \eqref{mg0.1} in the following form
\begin{align}
\label{mg}
\begin{cases}
\dot{x}_{i}(t)=-d_{i} x_{i}(t)-\Big[\nabla f\big(y(t^i_{k_i(t)})\big)\Big]_{i}+\theta_{i}\\[5pt]
y_{i}(t)=g_{i}\big(\lambda_ix_i(t)\big)\\[3pt]
\end{cases}
\end{align}
for $i=\oneton$. Here, $x_{i}\in\R$ , $d_{i}>0$ and $\theta_{i}\in\R$. In particular, let $d_{\max}=\max_{i}d_{i}$. $f(y):\R^{n}\to\R$ is an {\em analytic} cost function function and $y_{i}=g_{i}(\lambda_{i}x_{i})$ is the output vector with a scaling parameter $\lambda_{i}>0$ and the sigmoid functions $g_{i}(\cdot)$ as nonlinear activation functions, which we take as the sigmoid function as follows
\begin{align*}
g_{i}(x)=\frac{1}{1+e^{-x}}.
\end{align*}
The gradient of the activation function $g(\cdot)$ at $x\in\R^n$ can be written as $\partial g(x)=\textit{diag}\{g'_{1}(x_{1}),\cdots,g'_{n}(x_{n})\}$. The strict increasing triggering event time sequence $\{t_{k}^{i}\}_{k=1}^{+\infty}$ (to be defined) are
neuron-wise and $t_{1}^{i}=0$, for all $i=\oneton$. At each $t$, each neuron $i$ changes the information from the other neurons with respect to an identical time point $t_{k_{i}(t)}^{i}$ with $k_{i}(t)=\argmax_{k'}\{t^{i}_{k'}\leqslant t\}$. Throughout the paper, we simplify the notation $t_{k_{i}(t)}^{i}$ as $t_{k}^{i}$ unless there is a potential ambiguity.

Let $F(x_{t})=[F_{1}(x_{t}),\cdots,F_{n}(x_{t})]^{\top}$ be the vector at the right-hand side of \eqref{mg}, where
\begin{align}
F_{i}(x_{t})=-d_{i}x_{i}(t)-\Big[\nabla f\big(y(t^i_{k})\big)\Big]_{i}+\theta_{i}.\label{F}
\end{align}
Denote the set of equilibrium points for \eqref{mg} as
\begin{align*}
\mathcal S=\Big\{x\in\R^{n}:-Dx-\nabla f(g(\Lambda x))+\theta=0\Big\}.
\end{align*}
We first recall the definition of almost stability for model \eqref{mg} proposed in \cite{Mhi}.
\begin{definition}\label{convergence}
Given an analytic function $f(\cdot)$, the sigmoid function $g(\cdot)$ and constants $d_{i}$, $\theta_{i}$ and $\lambda_{i}$, system \eqref{mg} is said to be {\it almost sure stable} if for any initial values except a set of zero measure, the trajectory $x(t)$ of \eqref{mg}, there exists $x^{\bm*}\in\mathcal S$ such that
\begin{align*}
\lim_{t\to+\infty}x(t)=x^{\bm*}.
\end{align*}
\end{definition}
The following lemma shows that all solutions for \eqref{mg} are bounded and there exists at least one equilibrium point.
\begin{lemma}\label{Existence}
Given $d_{i}$, $\theta_{i}$, $\lambda_{i}$, $i=\oneton$, and two analytic functions $f(\cdot)$ and $g(\cdot)$, for any triggering event time sequence $\{t^{i}_k\}_{k=0}^{+\infty}~(i=\oneton)$, there exists a unique solution for the piece-wise cauchy problem \eqref{mg} with some initial data $x(0)\in\R^{n}$. Moreover, the solutions with different initial data are bounded in time interval of its duration.
\end{lemma}

\begin{IEEEproof} First, we prove the existence and uniqueness of the solution for the system (\ref{mg}). Denotes $t_{k}=[t^{1}_{k},\cdots,t^{n}_{k}]^{\top}$, where $k=\zerotoinfty$. Given a time sequence $\{t^{i}_k\}_{k=0}^{+\infty}~(i=\oneton)$ ordered as $0=t^{i}_{0}<t^{i}_{1}<t^{i}_{2}<\cdots<t^{i}_{k}<\cdots$ (same items in $\{t^{i}_k\}_{k=0}^{+\infty}$ treat as one), there exists a unique solution of \eqref{mg} in the interval $[t_{0},t_{1})$ by using $x(t_{0})=x(0)$ as the initial data according to the existence and uniqueness theorem in \cite{Jkh}). For the next interval $[t_{1},t_{2})$, $x(t_{1})$ can be regarded as the new initial data, which can derive another unique solution in this interval. By induction, we can conclude that there exists a piecewise unique solution over the whole time interval of the duration.

Second, since $0<g_i(x)<1~(x\in\R)$, there exists a constant $M>0$ such that
\begin{align*}
-d_{i}x_{i}(t)-M\leqslant F_{i}(x_{t})\leqslant-d_{i}x_{i}(t)+M
\end{align*}
Thus for any $\varepsilon_{0}>0$, there exists $r_0>0$ such that
\begin{align*}
\begin{cases}
F_{i}(x_{t})<-\varepsilon_0,&~\forall~ x_i(t)\geqslant r_0\\[2pt]
F_{i}(x_{t})>\varepsilon_0,&~\forall~ x_i(t)\leqslant -r_0
\end{cases}
\end{align*}
where $i=\oneton$. One can see that $x(t)\in\big\{x\in\R^{n}:\|x\|\leqslant \max\{r_{0},\|x(0)\|\}\big\}$ for the whole duration of the solution.
\end{IEEEproof}

Consider now the equilibrium points set $\mathcal S$. The following lemma is established in \cite{Mfa}, which shows that there exists at least one equilibrium point in $\mathcal S$.
\begin{lemma}\label{nontrivial}
For the equilibrium points set $\mathcal S$, the following statements hold:
\begin{enumerate}
\renewcommand{\labelenumi}{(\arabic{enumi})}
\item $\mathcal S$ is not empty.\\[-10pt]
\item There exists a constant $r>0$ such that $\mathcal S\bigcap\,\big(\R^{n}\setminus B_{r}(\bf0)\big)=\emptyset$.
\end{enumerate}
\end{lemma}

To depict the event that triggers the next feedback basing time point, we introduce the following candidate Lyapunov function:
\begin{align}
L(x)=&f(y)+\sum_{i=1}^{n}\bigg[\frac{d_{i}}{\lambda_{i}}\int_{0}^{y_{i}}g_{i}^{-1}(s){\rm d}s\bigg]
-\theta^{\top} y,\label{Ly}
\end{align}
where $y=[y_{1},\cdots,y_{n}]$ with $y_{i}=g_{i}(\lambda_{i}x_{i})~(i=\oneton)$. The function $L(x)$ generalizes the Lyapunov function introduced for \eqref{mg0.1} in \cite{Mvi}, and it can also be thought of as the energy function for the Hopfield and the cellular neural networks model \cite{Mac,Loc}. In this paper, we will prove that the candidate Lyapunov function \eqref{Ly} is a strict Lyapunov function \cite{Mfa}, defined as follows.
\begin{definition}
A Lyapunov function $L(\cdot):\R^{n}\rightarrow\R$ is said to be strict if $L\in C^{1}(\R^{n})$, and the derivative of $L$ along trajectories $x(t)$, i.e. $\dot{L}(x(t))$, satisfies $\dot{L}(x)\leqslant
0$  and  $\dot{L}(x)< 0$ for $x\notin \mathcal S$. 
\end{definition}

The next lemma provides the {\L}ojasiewicz inequality \cite{Slo}, which will be used to prove the
finiteness of length for any trajectory $x(t)$ of the system \eqref{mg}, towards convergence of the trajectory of the system \eqref{mg}.
\begin{lemma}\label{Loj}
Consider an analytic and continuous function $H(x):\mathcal D\subseteq\R^{n}\rightarrow\R$. Let
\begin{align*}
\mathcal S_{\nabla}=\Big\{x\in\mathcal D:\nabla H(x)=\bf 0\Big\}.
\end{align*}
For any $x_{s}\in\mathcal S_{\nabla}$, there exist two constants $r(x_{s})>0$ and $0<v(x_{s})<1$, such that
\begin{align*}
\big|H(x)-H(x_{s})\big|^{v(x_{s})}\leqslant\big\|\nabla H(x)\big\|,
\end{align*}
for $x\in B_{r(x_{s})}(x_{s})$.
\end{lemma}

The definition of trajectory length is given below.
\begin{definition}\label{def2}
Let $x(t)$ on $t\in[0,+\infty)$, be some trajectory of \eqref{mg}. For any
$t>0$, the length of the trajectory on $[0,t)$ is given by
\begin{align*}
l_{[0,t)}=\int_{0}^{t}\big\|\dot{x}(s)\big\|{\rm d}s.
\end{align*}
\end{definition}

\section{Event-trigger synaptic feedbacks and almost stability}\label{sec3}
In this section, we synthesize asynchronous triggers that prescribe when neurons should broadcast its state information and update their control signals.
Section \ref{Primary} presents the evolution of a quadratic function that measures network disagreement to identify a triggering function and discusses the problems that arise in its physical implementation. These observations are our starting point in Section \ref{ContinuousSituation} and Section \ref{AlternateSituation}, where we should overcome these implementation issues, also know as {\em Zeno behaviors}.

Define the state measurement error vector $e(t)=[e_{1}(t),\cdots,e_{n}(t)]^{\top}$ as
\begin{align*}
e_{i}(t)=\Big[\nabla f\big(y(t)\big)\Big]_{i}-\Big[\nabla f\big(y(t_{k_{i}(t)}^{i})\big)\Big]_{i}
\end{align*}
for $t\in[t_k^{i},t_{k+1}^{i})$ with $i=\oneton$ and $k=\zerotoinfty$.

\subsection{Event-trigger rule}\label{Primary}

For a constant $0<c<2$, denote
\begin{align*}
\alpha=\Big(1-\frac{c}{2}\Big)\inf_{i\in\oneton}\inf_{t\in[0,+\infty)}\Big\{\lambda_{i}g'\big(\lambda_{i}x_{i}(t)\big)\Big\}
\end{align*}
and
\begin{align*}
\beta=\frac{1}{2c}\sup_{i\in\oneton}\sup_{t\in[0,+\infty)}\Big\{\lambda_{i}g'\big(\lambda_{i}x_{i}(t)\big)\Big\}.
\end{align*}
The triggering function $T_{i}(e_{i},t)$ for the event-triggered rule can be defined as
\begin{align*}
T_{i}(e_{i},t)=\big|e_{i}(t)\big|-\gamma\,\Psi_{i}(t),
\end{align*}%
where $\gamma\in(0,\frac{\sqrt{\alpha}}{\sqrt{\beta}})$ and $\Psi_{i}(t)=\sqrt{\delta(t)}\,{\rm e}^{-d_{i}(t-t_{k_{i}(t)}^{i})}$
with
\begin{align}\label{normalization}
\delta(t)=\frac
{\sum\limits_{i=1}^{n}\big|F_{i}(x_{t})\big|^{2}}
{\sum\limits_{i=1}^{n}{\rm e}^{-2d_{i}(t-t_{k_{i}(t)}^{i})}}.
\end{align}
The updating rule for the trigger events is given as follow.
\begin{theorem}\label{PrimaryRule}
Given $T>0$, for each neuron $v_{i}$, set $t_{k+1}^{i}=\min\{\tau^{i}_{k+1},t^{i}_{k}+T\}$ with the following  updating rule:
\begin{align}\label{PrimaryRule1}
\tau_{k+1}^{i}=\max_{\tau\geqslant t_{k}^{i}}\bigg\{\tau:T_{i}(e_{i},t)
\leqslant 0,~\forall~ t\in[t^i_{k},\tau)\bigg\},
\end{align}
for all $i=\oneton$ and $k=\zerotoinfty$. Suppose that for a trajectory $x(t)$ of system (\ref{mg}), $\lim_{k\to\infty}t^{i}_{k}=+\infty$ holds for all $i$. Then $x(t)$ converges to some $x^{*}\in\mathcal S$, namely, $\lim_{t\to\infty}x(t)=x^{*}$.
\end{theorem}
Before the proof, we have the following remarks.
\begin{remark}
From (\ref{F}), one can see that it is sufficient to monitor  the neurons' states $x_{j}(t)$, $j=\oneton$, from system \eqref{mg}, in order to verify rule (\ref{PrimaryRule1}). And, the coefficient $\delta(t)$ in Eq. \eqref{normalization} can be seen as a parameter from this normalization process. Furthermore, according to (\ref{mg}), the evolution of the state ($x_{i}$) of neuron $v_{i}$ is in the form $\dot{x}_{i}=-d_{i}x_{i}+I_{i}$ with $I_{i}=\theta_{i}-[\nabla f(y(t_{k}^{i}))]_{i}$ a constant, before the event is triggered. Thus, we can either continuously monitoring neurons' states $x_{j}(t)$, $j=\oneton$, or formulate them as $x_{i}(t)=e^{-d_{i}(t-t^{i}_{k})}x_{i}(t^{i}_{k})+(1/d_{i})[1-e^{-d_{i}(t-t_{k}^{i})}]I_{i}$. Therefore, we can only monitor the states at the event times instead, which leads a discrete-time monitoring scheme. We will discuss this scenario in Section \ref{Monitoring} in detail.
\end{remark}
\begin{remark}
The preliminary condition for the convergence of the trajectory of system (\ref{mg}) is that the existing duration of the solution of the Cauchy problem of (\ref{mg}) should be $[0,\infty)$, or equivalently $\lim_{k\to\infty}t_{i}^{k}=\infty$ for all $i$. We will verify this condition by proving the exclusion of Zeno behaviors in Sections \ref{ContinuousSituation} and \ref{AlternateSituation}.
\end{remark}

The proof of this theorem comprises of the following propositions.

\begin{proposition}\label{Proposition1}
Under the assumptions in Theorem \ref{PrimaryRule}, $L(x)$ in \eqref{Ly} is a strict Lyapunov function for system \eqref{mg}.
\end{proposition}
\begin{IEEEproof}
The partial derivative of the candidate Lyapunov function $L(x)$ along the trajectory $x(t)$ can be written as
\begin{align}
\frac{\partial}{\partial x_i}L\big(x(t)\big)\nonumber
=&-\lambda_{i}g'\big(\lambda_{i}x_{i}(t)\big)\bigg\{-d_{i} x_{i}(t)
-\Big[\nabla f\big(y(t^{i}_{k_{i}(t)})\big)\Big]_{i}\nonumber\\
 &+\theta_{i}-\Big[\nabla f\big(y(t)\big)\Big]_{i}
+\Big[\nabla f\big(y(t^{i}_{k_{i}(t)})\big)\Big]_{i}\bigg\}\nonumber\\
=&-\lambda_{i}g'\big(\lambda_{i}x_{i}(t)\big)\Big[F_{i}(x_{t})-e_i(t)\Big],
\label{dLyx}
\end{align}
and the time derivative of $L(x(t))$ gives
\begin{align*}
\dot{L}\big(x(t)\big)=&\sum_{i=1}^{n}\frac{\partial}{\partial x_{i}}L\big(x(t)\big)\frac{{\rm d}x_{i}(t)}{{\rm d}t}\nonumber\\
=&-\sum_{i=1}^{n}\lambda_{i}g'\big(\lambda_{i}x_{i}(t)\big)\Big[F_{i}(x_{t})-e_{i}(t)\Big]F_{i}(x_{t}).
\end{align*}
Consider the inequality $|e_{i}(t)F_{i}(x_{t})|\leqslant\frac{1}{2c}|e_{i}(t)|^{2}+\frac{c}{2}|F_{i}(x_{t})|^{2}$ for some $0<c<2$. Then we have
\begin{align*}
 \dot{L}\big(x(t)\big)
=&-\sum_{i=1}^{n}\lambda_{i}g'\big(\lambda_{i}x_{i}(t)\big)
\bigg[\big|F_{i}(x_{t})\big|^{2}-e_{i}(t)F_{i}(x_{t})\bigg]\nonumber\\
\leqslant
 &-\Big(1-\frac{c}{2}\Big)\sum_{i=1}^{n}\lambda_{i}g'\big(\lambda_{i}x_{i}(t)\big)
 \big|F_{i}(x_{t})\big|^{2}\nonumber\\
 &+\frac{1}{2c}\sum_{i=1}^{n}\lambda_{i}g'\big(\lambda_{i}x_{i}(t)\big)\big|e_{i}(t)\big|^{2}\nonumber\\
\leqslant
 &-\alpha\sum_{i=1}^{n}\big|F_{i}(x_{t})\big|^{2}+\beta\sum_{i=1}^{n}\big|e_{i}(t)\big|^{2}
\end{align*}
From the rule \eqref{PrimaryRule1}, one can see that $|e_{i}(t_{k}^{i}+T)|\leqslant\gamma\Psi_{i}(t_{k}^{i}+T)$  holds for all $t\geqslant 0$ at most except $t=t^{i}_{1},\cdots,t^{i}_{k},\cdots$. This implies
\begin{align}\label{dLy0.1}
\dot{L}\big(x(t)\big)
\leqslant
&-\alpha\sum_{i=1}^{n}\big|F_{i}(x_{t})\big|^{2}
+\beta\gamma^{2}\sum_{i=1}^{n}\Psi_{i}^{2}(t)\nonumber\\
=&-\alpha\sum_{i=1}^{n}\big|F_{i}(x_{t})\big|^{2}
  +\beta\gamma^{2}\sum_{i=1}^{n}\delta(t)\,{\rm e}^{-2d_{i}(t-t_{k_{i}(t)}^{i})}\nonumber\\
=&-\big(\alpha-\beta\gamma^{2}\big)\sum_{i=1}^{n}\big|F_{i}(x_{t})\big|^{2}\leqslant0
\end{align}
for all $k=\zerotoinfty$. For any $x\notin \mathcal S$, there exits $i_0\in\{\oneton\}$ such that
$F_{i_0}(x_{t})\neq0$. Thus $\dot{L}(x)<0$. Proposition \ref{Proposition1} is proved.
\end{IEEEproof}

With the Lyapunov function $L(x)$ for system (\ref{mg}) and the event triggering condition \eqref{PrimaryRule1}, the consequent proof follows \cite{Mfa} with necessary modifications.
\begin{proposition}\label{Proposition2}
There exist finite different energy levels $L_j~(j=\onetom)$, such that each set
of equilibrium points
\begin{align*}
\mathcal S_j=\Big\{x\in\mathcal S:L(x)=L_j\text{ and }j=\onetom\Big\}
\end{align*}
is not empty.
\end{proposition}
\begin{IEEEproof}
First of all, it can be seen that $L(x)$ in \eqref{Ly} is analytic on $\R^{n}$.
Suppose that there exist infinite different values
$L_{j}~(j=1,\cdots,+\infty)$ such that $\mathcal S_{j}=\{x\in\mathcal
S:L(x)=L_{j}\}$ is not empty. From Lemma \ref{nontrivial}, it is known that
there exists $r_{1}>0$ such that outside $B_{r_{1}}(\bf 0)$ there are no
equilibrium points. Hence $\mathcal S_{j}\subset B_{r_{1}}(\bf 0)$ for
$j=1,\cdots,+\infty$.

Consider points $x^{j}\in\mathcal S_{j}$ for $j=1,\cdots,+\infty$. Since
$x^{j}\in\mathcal S$, it holds $F(x^{j})=0$ and from Eq. \eqref{dLyx}, $\nabla
L(x^{j})=\bf 0$. Since $\overline{B_{r_{1}}(\bf 0)}$ is a compact set, hence, there
exist a point $\widetilde{x}$ and a subsequence
$\{x^{j_{h}}\}_{h=1}^{+\infty}$ such that $x^{j_{h}}\neq\widetilde{x}$ for
all $h=1,\cdots,+\infty$ and $x^{j_{h}}\to\widetilde{x}$ as $h\to+\infty$.
Since $\nabla L$ is continuous, taking into account that $\nabla
L(x^{j_{h}})=\bf 0$ for all $h=1,\cdots,+\infty$, it results $\nabla
L(\widetilde{x})=\bf 0$.

According to Lemma \ref{Loj}, there exist $r(\widetilde{x})>0$ and
$v(\widetilde{x})\in(0,1)$ such that
$|L(x)-L(\widetilde{x})|^{v(\widetilde{x})}\leqslant\|\nabla
L(x)\|$ for $x\in B_{r(\widetilde{x})}(\widetilde{x})$. Since
$x^{j_{h}}\to\widetilde{x}$ as $h\to+\infty$ and $x^{j_{h}}\in\mathcal
S_{j_{h}}$ have different energy levels $L_{j_{h}}$, we can pick a point
$x^{j_{h_{0}}}\in B_{r(\widetilde{x})}(\widetilde{x})$ such that
$L(x^{j_{h_{0}}})\neq L(\widetilde{x})$. Then
\begin{align*}
0<\big|L(x^{j_{h_{0}}})-L(\widetilde{x})\big|^{v(\widetilde{x})}
\leqslant\big\|\nabla L(x^{j_{h_{0}}})\big\|=0,
\end{align*}
which is a contradiction. This completes the proof.
\end{IEEEproof}

Without loss of generality, assume that the energy levels
$L_{j}~(j=\onetom)$ are ordered as $L_{1}>L_{2}>\cdots>L_{m}$. Thus
there exists $\gamma>0$ such that $L_{j}>L_{j+1}+2\gamma$, for any
$j=1,2,\cdots,m-1$. For any given $\varepsilon>0$, define
\begin{align*}
\Gamma_{j}=\Big\{x\in\R^{n}:\text{dist}\,(x,\mathcal S_{j})\leqslant\varepsilon\Big\}
\end{align*}
and
\begin{align}\label{Kset}
\mathcal
K_{j}=\overline{\Gamma}_{j}\bigcap\Big\{x\in\R^{n}:L(x)\in\big[L_{j}-\gamma,L_{j}+\gamma\big]\Big\}.
\end{align}

\begin{proposition}\label{Proposition3}
For $j=1,2,\cdots,m$, $\mathcal K_{j}$ is a
compact set and $\mathcal K_{j}\bigcap\mathcal S=\mathcal S_{j}$.
\end{proposition}
\begin{IEEEproof} From Lemma \ref{nontrivial}, $\mathcal S_{j}\in B_{r_{1}}(\bf
0)$ is bounded, hence $\overline{\Gamma}_{j}$ is a compact set and
$\big\{x\in\R^{n}:L(x)\in[L_{j}-\gamma,L_{j}+\gamma]\big\}$ is a closed
set. Thus, $\mathcal
K_{j}=\overline{\Gamma}_{j}\bigcap\big\{x\in\R^{n}:L(x)\in[L_{j}-\gamma,L_{j}+\gamma]\big\}$
is a compact set. Then proterty $\mathcal K_{j}\bigcap\mathcal S=\mathcal
S_{j}$ is an immediate consequence of Proposition \ref{Proposition2}.
\end{IEEEproof}

\begin{proposition}\label{Proposition4}
For any trajectory $x(t)$ of the system \eqref{mg} and any given time point $\tau\geqslant0$, let $\mathcal K_{j}$, for some
$j\in\{\onetom\}$, be a compact set as defined in \eqref{Kset}. Then there exist a constant $C_{j}>0$ and an exponent
$v_{j}\in(0,1)$ such that
\begin{align*}
\frac{\Big|\dot{L}\big(x(\tau)\big)\Big|}{\big\|F(x_{\tau})\big\|}
\geqslant C_{j}\Big|L\big(x(\tau)\big)-L_{j}\Big|^{v_{j}},
\end{align*}
for $x(\tau)\in\mathcal K_{j}\setminus \mathcal S$.
\end{proposition}

\begin{IEEEproof}
Recall the notion $t_{k_{i}(\tau)}^{i}$  as $t_{k}^{i}$ with $k_{i}(\tau)=\arg\max_{k'}\{t^{i}_{k'}\leqslant\tau\}$. $F_{i}(x_{t})$ can be rewritten as
\begin{align*}
F_{i}(x_{\tau})=-d_{i}x_{i}(t_{k}^{i})-\Big[\nabla f\big(y(t_{k}^{i})\big)\Big]_{i}+\theta_{i},
\end{align*}
for $i=\oneton$. From \eqref{dLyx} and \eqref{PrimaryRule1}, we have
\begin{align*}
 &~\Big\|\nabla L\big(x(\tau)\big)\Big\|^{2}\\
=&\sum_{i=1}^{n}\bigg|\frac{\partial}{\partial x_i}L\big(x(\tau)\big)\bigg|^{2}\\
=&\sum_{i=1}^{n}\bigg|\lambda_{i}g'\big(\lambda_{i}x_{i}(\tau)\big)\Big[F_{i}(x_{\tau})-e_{i}(\tau)\Big]\bigg|^{2}\\
\leqslant
 &~\tilde{\beta}_{j}^{2}\sum_{i=1}^{n}
\Bigg[F_{i}^{2}(x_{\tau})+e_{i}^{2}\big(x(\tau)\big)+2\Big|F_{i}(x_{\tau})e_{i}\big(x(\tau)\big)\Big|\Bigg]\\
\leqslant
 &~\tilde{\beta}_{j}^{2}\sum_{i=1}^{n}
  \Bigg[(1+c)F_{i}^{2}(x_{\tau})+\bigg(1+\frac{1}{c}\bigg)e_{i}^{2}\big(x(\tau)\big)\Bigg]\\
\leqslant
 &~\tilde{\beta}_{j}^{2}(1+c)\sum_{i=1}^{n}\big|F_{i}(x_{\tau})\big|^{2}+
  ~\tilde{\beta}_{j}^{2}\bigg(1+\frac{1}{c}\bigg)\gamma^{2}\sum_{i=1}^{n}\Psi_{i}^{2}(\tau)\\
=&~\tilde{\beta}_{j}^{2}(1+c)\bigg(1+\frac{\gamma^{2}}{c}\bigg)\big\|F(x_{\tau})\big\|^{2},
\end{align*}
where $\tilde{\beta}_{j}=\max_{i\in\{\oneton\}}\max_{x(\tau)\in\mathcal K_{j}}\{\lambda_{i}g'_{i}(\lambda_{i}x_{i}(\tau))\}$.
Then it holds
\begin{align*}
\big\|F(x_{\tau})\big\|\geqslant h_{j}\Big\|\nabla L\big(x(\tau)\big)\Big\|,
\end{align*}
where
\begin{align*}
h_{j}=\frac{1}{\tilde{\beta}_{j}\sqrt{(1+c)\big(1+\frac{\gamma^{2}}{c}\big)}}.
\end{align*}
From Eq. \eqref{dLy0.1}, we have
\begin{align*}
\Big|\dot{L}\big(x(\tau)\big)\Big|\geqslant\big(\alpha-\beta\gamma^{2}\big)\big\|F(x_{\tau})\big\|^{2}.
\end{align*}

For the point $x(\tau)\in\mathcal K_{j}\setminus \mathcal S$, from Eq. \eqref{dLyx},
$\nabla L(x(\tau))\neq0$. There exists
$r(x(\tau))>0$, $c(x(\tau))>0$ and an exponent
$v(x(\tau))\in(0,1)$ such that
\begin{align*}
\Big\|\nabla L\big(x(\tau)\big)\Big\|\geqslant c\big(x(\tau)\big)\Big|L\big(x(\tau)\big)-L_{j}\Big|^{v(x(\tau))},
\end{align*}
for $x\in B_{r(x(\tau))}(x(\tau))$. Indeed, if
$r(x(\tau))>0$ is small, we have $\nabla L(x)\neq0$ for
$x\in\overline{B_{r(x(\tau))}(x(\tau))}$. Therefore, it holds
\begin{align*}
\frac{\Big|\dot{L}\big(x(\tau)\big)\Big|}{\big\|F(x_{\tau})\big\|}
&\geqslant\big(\alpha-\beta\gamma^{2}\big)\big\|F(x_{\tau})\big\|\\
&\geqslant\big(\alpha-\beta\gamma^{2}\big)h_{j}\Big\|\nabla L\big(x(\tau)\big)\Big\|\\[3pt]
&\geqslant\big(\alpha-\beta\gamma^{2}\big)h_{j}c\big(x(\tau)\big)\Big|L\big(x(\tau)\big)-L_{j}\Big|^{v(x(\tau))}\\[3pt]
&\geqslant~C_{j}\Big|L\big(x(\tau)\big)-L_{j}\Big|^{v_{j}},
\end{align*}
where
\begin{align*}
C_{j}=(\alpha-\beta\gamma^{2})\,h_{j}\min_{x(\tau)\in\mathcal K_{j}}\Big\{c\big(x(\tau)\big)\Big\}
\end{align*}
and
\begin{align*}
v_{j}=\min_{x(\tau)\in\mathcal K_{j}}\Big\{v\big(x(\tau)\big)\Big\}
\end{align*}
for $x(\tau)\in \mathcal K_{i}\setminus \mathcal S$.
\end{IEEEproof}

Now, we are at the stage to prove that the length of $x(t)$ on $[0,+\infty)$ is finite. The statement proposition is given as follow.
\begin{proposition}\label{Proposition5}
Any trajectory $x(t)$ of the systm \eqref{mg} has a finite length on $[0,+\infty)$, i.e.,
\begin{align*}
l_{[0,+\infty)} =\int_{0}^{+\infty}\big\|\dot{x}(s)\big\|{\rm d}s
=\lim_{t\rightarrow+\infty}\int_{0}^{t}\big\|\dot{x}(s)\big\|{\rm d}s<+\infty.\\[-10pt]
\end{align*}
\end{proposition}

\begin{IEEEproof}
Assume without loss of generality that $x(0)$ is not an equilibrium point
of Eq. \eqref{mg}. Due to the uniqueness of solutions, we have
$\dot{x}(t)=F(x_{t})\neq 0$ for $t\geqslant 0$, i.e.,
$x(t)\in\R^{n}\setminus \mathcal S$ for $t\geqslant 0$. From Proposition \ref{Proposition1}, it is seen that $L(x(t))$ satisfies
$\dot{L}(x(t))<0$ for $t\geqslant0$, i.e., $L(x(t))$
strictly decreases for $t\geqslant0$. Thus, since $x(t)$ is bounded on
$[0,+\infty)$ and $L(x(t))$ is continuous, $L(x(t))$ will
tend to a finite value $L(+\infty)=\lim_{t\to+\infty}L(x(t))$. From
Proposition \ref{Proposition1} and the LaSalle invariance principle \cite{Mhi},
\cite{Jkh}, it also follows that $x(t)\to\mathcal S~(t\to+\infty)$. Thus,
from the continuity of $L$, it results $L(+\infty)=L_{j}$ for some
$j\in\{\onetom\}$ and $x(t)\to \mathcal S_{j}~(t\to+\infty)$.

Since $x(t)\to \mathcal S_{j}~(t\to+\infty)$ and $L(x(t))\to
L_{j}~(t\to+\infty)$, it follows that there exists $\widetilde{t}>0$ such
that $x(t)\in\mathcal K_{i}$ for $t\geqslant\widetilde{t}$.
By using Proposition \ref{Proposition4}, considering that
$x(t)\in\R^{n}\setminus\mathcal S$ for $t\geqslant 0$ and $x(t)\in\mathcal
K_{i}$ for $t\geqslant\widetilde{t}$, we have that there exists $C_{j}>0$
and $v_{j}\in(0,1)$ such that
\begin{align*}
\frac{\Big|\dot{L}\big(x(t)\big)\Big|}{\big\|F(x_{t})\big\|}
=\frac{-\dot{L}\big(x(t)\big)}{\big\|F(x_{t})\big\|}
\geqslant C_{j}\Big|L\big(x(t)\big)-L(+\infty)\Big|^{v_{j}},
\end{align*}
for $t\geqslant\widetilde{t}$. Then
\begin{align*}
\int_{\widetilde{t}}^{t}\big\|\dot{x}(s)\big\|{\rm d}s
&=\int_{\widetilde{t}}^{t}\big\|F(x_{s})\big\|{\rm d}s\\
&\leqslant\frac{1}{C_{j}}
\int_{\widetilde{t}}^{t}\frac{-\dot{L}\big(x(s)\big)}{\Big|L\big(x(s)\big)-L(+\infty)\Big|^{v_{j}}}ds.
\end{align*}
The change of variable $\sigma=L(x(s))$ derives
\begin{align*}
\int_{\widetilde{t}}^{t}\big\|\dot{x}(s)\big\|{\rm d}s
\leqslant&\,\frac{1}{C_{j}}\int_{L(x(\widetilde{t}))}^{L(x(t))}-\frac{1}{\big|\sigma-L(+\infty)\big|^{v_{j}}}d\sigma\\
=&\,\frac{1}{C_{j}(1-v_{j})}\Bigg\{\Big[L\big(x(\widetilde{t})\big)-L(+\infty)\Big]^{1-v_{j}}\\
 &-\Big[L\big(x(t)\big)-L(+\infty)\Big]^{1-v_{j}}\Bigg\}\\
\leqslant&\,\frac{1}{C_{j}(1-v_{j})}\Big[L\big(x(\widetilde{t})\big)-L(+\infty)\Big]^{1-v_{j}},
\end{align*}
for $t\geqslant\widetilde{t}$. Therefore, we have
\begin{align*}
l_{[0,+\infty)}
&=\int_{0}^{+\infty}\big\|\dot{x}(s)\big\|{\rm d}s\\
&\leqslant\int_{0}^{\widetilde{t}}\big\|\dot{x}(s)\big\|{\rm d}s
+\int_{\widetilde{t}}^{+\infty}\big\|\dot{x}(s)\big\|{\rm d}s\\
&\leqslant\int_{0}^{\widetilde{t}}\big\|\dot{x}(s)\big\|{\rm d}s
+\frac{\Big[L\big(x(\widetilde{t})\big)-L(+\infty)\Big]^{1-v_{j}}}{C_{j}(1-v_{j})}\\
&<+\infty.
\end{align*}
This completes the proof of Proposition \ref{Proposition5}.
\end{IEEEproof}

In what follows it remains to address the proof of Theorem \ref{PrimaryRule}, which is given in Section \ref{Primary}.

\textit{Proof of Theorem \ref{PrimaryRule}:}
Suppose that the condition \eqref{PrimaryRule1} holds. Then from Proposition \ref{Proposition5}, for any trajectory $x(t)$ of the system \eqref{mg}, we have
\begin{align*}
l_{[0,+\infty)}
=\int_{0}^{+\infty}\big\|\dot{x}(s)\big\|{\rm d}s
=\lim_{t\rightarrow+\infty}\int_{0}^{t}\big\|\dot{x}(s)\big\|{\rm d}s<+\infty.
\end{align*}
From Cauchy criterion on limit existence, for any $\varepsilon>0$,
there exists $T(\varepsilon)$ such that when $t_{2}>t_{1}>T(\varepsilon)$, it results
$\int_{t_{1}}^{t_{2}}\big\|\dot{x}(s)\big\|ds<\varepsilon$. Thus,
\begin{align*}
\big\|x(t_{1})-x(t_{2})\big\|
=\bigg\|\int_{t_{1}}^{t_{2}}\dot{x}(s){\rm d}s\bigg\|
\leqslant\int_{t_{1}}^{t_{2}}\big\|\dot{x}(s)\big\|{\rm d}s
<\varepsilon.
\end{align*}
It follows that there exists an equilibrium point $x^{\bm*}$ of \eqref{mg}, such that $\lim_{t\rightarrow+\infty}x(t)=x^{\bm*}$. Recalling the Definition \ref{convergence}, we can conclude that $x(t)$ is convergent.{\hspace*{\fill}\IEEEQED}

\begin{remark}
The event-triggered condition \eqref{PrimaryRule1} implies that the next time interval for neuron $v_{i}$ depends on states of the neurons $v_{j}$ that are synaptically linked to neuron $v_{i}$
We say that neuron $v_{j}$ is synaptically linked to neuron $v_{i}$ if $[\nabla f(y)]_{i}$ depends on $y_{j}$, in other words,
\begin{align*}
\frac{\partial^{2} f(y)}{\partial y_{i}\partial y_{j}}\ne 0.
\end{align*}
\end{remark}

When the event triggers, the neuron $v_{i}$ has to send its current state information $x_{i}(t)$ to the other neurons immediately in order to avoid having $\frac{\rm d}{{\rm d}t}L(x(t))>0$. However, such a trigger rule would cause the following problems:
\begin{enumerate}
\renewcommand{\labelenumi}{(P\arabic{enumi})}
\item\label{P1} The triggering function $T_{i}(e_{i},t)=0$ may hold even after neuron $v_{i}$ sends its new state to the other neurons. A bad situation is that $\Psi_{i}(t)=0$ happens at the same time when $|e_{i}(t)|=0$. This may cause the neuron to send its state continuously. This is called {\it continuous triggering situation} in the Zeno behavior \footnote{~Zeno behavior is described as a system making an infinite number of jumps (i.e. triggering events in this paper) in a finite amount of time (i.e. a finite time interval in this paper), see \cite{Khj}.}.
\item\label{P2} Event if $\Psi_{i}(t)=0$ and $|e_{i}(t)|=0$ never happen at the same time point. The Zeno behavior may still exist. For example, one neuron $v_{i}$ broadcasting its new state to the other neurons may cause the triggering rules for two neurons $v_{j_{1}}$ and $v_{j_{2}}$ are broken alternately. That is to say, the inter-event time for both $v_{j_{1}}$ and $v_{j_{2}}$ will decrease to zero. This is called {\it alternate triggering situation} in the Zeno behavior.
\end{enumerate}
These observations motivate us to introduce the Morse-Sard Theorem for avoiding the {\it continuous triggering situation} (P\ref{P1}) in Subsection \ref{ContinuousSituation}. In Subsection \ref{AlternateSituation}, we will also prove that for all the neuron, the {\it alternate triggering situation} is absent by the event-triggered rule in Theorem \ref{PrimaryRule}.

\subsection{Exclusion of continuous triggering situation }\label{ContinuousSituation}
From the rule \eqref{PrimaryRule1}, we know that a triggering event happens at a threshold time $t_{k}^{i}$ satisfying
\begin{align*}
T_{i}(e_{i},t_{k}^{i})=\big|e_{i}(t_{k}^{i})\big|-\gamma\Psi_{i}(t_{k}^{i})=0
\end{align*}
for $i=\oneton$ and $k=\zerotoinfty$.

To avoid the situation that $\Psi_{i}(t)=0$ and $|e_{i}(t)|=0$ happen at the same triggering time point $t_{k}^{i}$ for some $k$, when the triggering function $T_{i}(e_{i},t)=0$ still holds after the neuron $v_{i}$ sends the new state to the other neurons, we define a function vector
\begin{align*}
S(t,t_{\tau})=\frac{1}{2}\Big[e^{\top}(t)e(t)-\gamma^{2}\Psi^{\top}(t)\Psi(t)\Big]
\end{align*}
where $t_{\tau}\in\mathcal T_{k}$ and
\begin{align}\label{ThresholdTime}
\mathcal T_{k}=\bigcup_{i=1}^{n}\Big\{t^{i}_k:t_{k}^{i}\leqslant t\Big\} 
\end{align}
for $k=\zerotoinfty$, which is the set of all the latest triggering time points before the present time $t$.
Denote $S(t,t_{\tau})=\big[S_{1}(t,t_{\tau}),\cdots,S_{n}(t,t_{\tau})\big]^{\top}$ and the following theorem that comes from the Morse-Sard theorem \cite{Apm,As} will be used for excluding this continuous triggering.

\begin{theorem}\label{MSTheorem}
For each initial data $x(0)$, there exists a measure zero subset $\mathcal O\subset\R^{n}$ such that for all the neurons $v_{i}~(i=\oneton)$, if $x(0)\in\R^{n}\backslash\mathcal O$ and the triggering time point set at the first event $\mathcal T_{0}$ is countable, then the set of all the triggering time point on $[0,+\infty)$ 
\begin{align*}
\mathcal T=\bigcup_{k=0}^{+\infty}\bigcup_{i=1}^{n}\big\{t_{k}^{i}\big\}
\end{align*}
is also a countable set.
\end{theorem}

\begin{IEEEproof}
To show that the triggering time point set $\mathcal T$ is countable for each $x(0)\in\R^{n}\backslash\mathcal O$, we first prove an equivalent statement that the Jacobian matrix ${\rm d}S(t,t_{\tau})=\big[{\rm d}S_{1}(t,t_{\tau}),\cdots,{\rm d}S_{n}(t,t_{\tau})\big]^{\top}$ has rank $n$ at next triggering time point $t=t_{k+1}^{i}$ when $\mathcal T_{k}$ (i.e., the set of all the latest triggering time points before the present time $t$) is countable.

Note
\begin{align*}
{\rm d}S_{i}(t,t_{\tau})=
\bigg[
\frac{\partial}{\partial t}S_{i}(t,t_{\tau}),
\frac{\partial}{\partial t_{\tau}}S_{i}(t,t_{\tau})
\bigg],
\end{align*}
where $t_{\tau}\in \mathcal T_{k}$. The two components in the above equation satisfy
\begin{align*}
 \frac{\partial}{\partial t}S_{i}(t,t_{\tau})
=&~e_{i}(t)\frac{{\rm d}e_{i}(t)}{{\rm d}t}-\gamma^{2}\Psi_{i}(t)\frac{{\rm d}\Psi_{i}(t)}{{\rm d}t}\\
=&~e_{i}(t)\frac{{\rm d}}{{\rm d}t}\Big[\nabla f\big(y(t)\big)\Big]_{i}
  +\gamma^{2}d_{i}\delta^{2}(t)\,{\rm e}^{-2d_{i}(t-t_{k}^{i})}\\
=&~e_{i}(t)\frac{{\rm d}}{{\rm d}t}\Big[\nabla f\big(y(t)\big)\Big]_{i}
  +\gamma^{2}d_{i}\Psi_{i}^{2}(t)
\end{align*}
and
\begin{align*}
  \frac{\partial}{\partial t_{\tau}}S_{i}(t,t_{\tau})
=&~e_{i}(t)\frac{{\rm d}e_{i}(t)}{{\rm d}t_{\tau}}
  -\gamma^{2}\Psi_{i}(t)\frac{{\rm d}\Psi_{i}(t)}{{\rm d}t_{\tau}}\\
=&-e_{i}(t)\frac{{\rm d}}{{\rm d}t_{\tau}}\Big[\nabla f\big(y(t_{\tau})\big)\Big]_{i}
  -\gamma^{2}d_{i}\delta^{2}(t)\,{\rm e}^{-2d_{i}(t-t_{k}^{i})}\\
=&-e_{i}(t)\frac{{\rm d}}{{\rm d}t_{\tau}}\Big[\nabla f\big(y(t_{\tau})\big)\Big]_{i}
  -\gamma^{2}d_{i}\Psi_{i}^{2}(t)
\end{align*}
When event triggers and $e_{i}(t)$ resets to $0$ in the short time period after the next time point $t_{k+1}^{i}$, that is, $e_{i}(t_{k+1}^{i}+\varepsilon)\to0$ when $\varepsilon\to0$, then it holds
\begin{align*}
 \lim_{\varepsilon\to0}
 \frac{\partial}{\partial t}S_{i}(t,t_{\tau})\bigg|_{t=t_{k+1}^{i}+\varepsilon}
=\gamma^{2}d_{i}\Psi_{i}^{2}(t_{k+1}^{i})
\end{align*}
and
\begin{align*}
  \lim_{\varepsilon\to0}
  \frac{\partial}{\partial t_{\tau}}S_{i}(t,t_{\tau})\bigg|_{t=t_{k+1}^{i}+\varepsilon}
=-\gamma^{2}d_{i}\Psi_{i}^{2}(t_{k+1}^{i}).
\end{align*}
Define a initial data set by 
\begin{align*}
\mathcal O_{k}^{i}=\bigg\{x_{i}(0)\in\R:\delta\big(t_{k+1}^{i}\big)=0\text{~for all~}i=\oneton\bigg\}
\end{align*}
 Take the initial data $x(0)\in\R^{n}\backslash\bigcup_{i=1}^{n}\mathcal  O_{k}^{i}$, we have
\begin{align*}
\Psi_{i}(t_{k+1}^{i})=\sqrt{\delta(t_{k+1}^{i})}\,{\rm e}^{-d_{i}(t_{k+1}^{i}-t_{k}^{i})}\neq0
\end{align*}
which implies
\begin{align*}
{\rm d}S_{i}\big(t_{k+1}^{i},t_{\tau}\big)
=&~\lim_{\varepsilon\to0}{\rm d}S_{i}\big(t_{k+1}^{i}+\varepsilon,t_{\tau}\big)\\
=&~\lim_{\varepsilon\to0}\bigg[
  \frac{\partial}{\partial t}S_{i}\big(t,t_{\tau}\big),
  \frac{\partial}{\partial t_{\tau}}S_{i}\big(t,t_{\tau}\big)
  \bigg]\bigg|_{t=t_{k+1}^{i}+\varepsilon}\\
=&~\Big[\gamma^{2}d_{i}\Psi_{i}^{2}(t_{k+1}^{i}),-\gamma^{2}d_{i}\Psi_{i}^{2}(t_{k+1}^{i})\Big]\\[2pt]
\neq&~0.
\end{align*}
that is, the Jacobian matrix ${\rm d}S(t,t_{\tau})$ has rank $n$ at $t=t_{k+1}^{i}$.

Then it follows that if $x(0)\in\R^{n}\backslash\bigcup_{i=1}^{n}\mathcal  O_{k}^{i}$ and $\mathcal T_{k}$ is countable, the next triggering time point $t_{k+1}^{i}$ is isolated, hence the next triggering time point set for all the neurons
\begin{align*}
\mathcal T_{k+1}=\bigcup_{i=1}^{n}\big\{t_{k+1}^{i}\big\}
\end{align*}
is countable. Furthermore, by using the inverse function theorem, it holds
\begin{align*}
\measure(O_{k})=0
\text{~and~}
\mathcal O_{k}=\bigcup_{i=1}^{n}\mathcal O_{k}^{i}
\end{align*}
for the Lebesgue measure $\measure(\cdot)$. 

Now, according to the method of induction, we can assert that for each initial data $x(0)\in\R^{n}\backslash\mathcal O$, where
\begin{align}\label{ZeroMeasuredSet}
\mathcal O=\bigcup_{k=0}^{+\infty}\mathcal O_{k}=\bigcup_{k=0}^{+\infty}\bigcup_{i=1}^{n}\mathcal O_{k}^{i},
\end{align}
the triggering time points set for all the neuron on $[0,+\infty)$
\begin{align*}
\mathcal T=\bigcup_{k=0}^{+\infty}\mathcal T_{k}=\bigcup_{k=0}^{+\infty}\bigcup_{i=1}^{n}\big\{t_{k}^{i}\big\}
\end{align*}
is countable and moreover $\measure(O)=0$ under the assumption that $\mathcal T_{0}$ is countable. This theorem is proved.
\end{IEEEproof}

Recalling the triggering function $T_{i}(e_{i},t)$, we have the results that if the initial data $x(0)\in\R^{n}\backslash\mathcal O$ and there exist countable number of triggering time points at the first event (i.e., $\mathcal T_{0}$ is countable), then
\begin{align*}
\Psi(t_{\tau})\neq0
\text{~ for ~} 
t_{\tau}\in\mathcal T=\bigcup_{k=0}^{+\infty}\bigcup_{i=1}^{n}\big\{t_{k}^{i}\big\}
\end{align*}
that is to say, $\Psi_{i}(t)=0$ and $|e_{i}(t)|=0$ may never happen at the same time at all the triggering time point $t_{k}^{i}$ where $i=\oneton$ and $k=\zerotoinfty$. Therefore, the {\it continuous triggering situation} in the Zeno behavior (P\ref{P1}) is avoided by the rule \eqref{PrimaryRule1}.
\begin{remark}
Any perturbation on the initial data $x(0)$ can help away from the zero measured subset $\mathcal O$.
\end{remark}

\subsection{Exclusion of alternate triggering situation}\label{AlternateSituation}
After excluding the {\it continuous triggering situation}, we are to prove the absence of the {\it alternate triggering situation} in the Zeno behavior. Toward this aim, we will find a common positive lower-bound for the inter-event time $t_{k+1}^{i}-t_{k}^{i}$, for all $i=\oneton$ and $k=\zerotoinfty$.
\begin{theorem}\label{Zeno}
Let $\mathcal O$ be a zero measured set as defined in \eqref{ZeroMeasuredSet}. Under two criterions of the event-triggered rule in Theorem \ref{PrimaryRule}, for each $x(0)\in\R^{n}\backslash\mathcal O$, the next inter-event interval of every neuron is strictly positive and has a common positive lower-bound. Furthermore,
the {\it alternate triggering situation} in the Zeno behaviors are excluded.
\end{theorem}

\begin{IEEEproof}
Let us consider the following derivative of the state measurement error for any neuron $v_{i}~(i=\oneton)$
\begin{align*}
 ~\big|\dot{e}_{i}(t)\big|
=&~\Bigg|\sum_{j=1}^{n}\Big[\nabla^2 f\big(y(t)\big)\Big]_{ij}\,\dot{y_{j}}(t)\Bigg|\\[2pt]
=&~\Bigg|\sum_{j=1}^{n}\Big[\nabla^2 f\big(y(t)\big)\Big]_{ij}
   \lambda_{j}g'_{j}\big(\lambda_{j}x_{j}(t)\big)F_{j}\big(x_{t}\big)\Bigg|\\
\leqslant
 &~\Big\|\nabla^{2} f\big(y(t)\big)\Big\|
   \Big\|\Lambda\,\partial g\big(\Lambda x(t)\big)\Big\|
   \sqrt{\sum_{j=1}^{n}\Big|F_{j}\big(x_{t}\big)\Big|^{2}}\\
=
 &~\Big\|\nabla^{2} f\big(y(t)\big)\Big\|\Big\|\Lambda\Big\|
   \sqrt{\delta(t)\sum\limits_{i=1}^{n}{\rm e}^{-2d_{i}(t-t_{k}^{i})}}\\
\leqslant
 &~\mathcal M\sqrt{\delta\big(t\big)}
\end{align*}
where
\begin{align*}
\mathcal M=\sqrt{n}\big\|\Lambda\big\|\sup_{t\in[0,+\infty)}\Big\|\nabla^{2} f\big(y(t)\big)\Big\|
\end{align*}
and then it follows
\begin{align*}
\big|e_{i}(t)\big|
=\bigg|\int_{t_{k}^{i}}^{t}\dot{e}_{i}(s){\rm d}s\bigg|
\leqslant\int_{t_{k}^{i}}^{t}\big|\dot{e}_{i}(s)\big|{\rm d}s
\leqslant\mathcal M\int_{t_{k}^{i}}^{t}\sqrt{\delta(s)}{\rm d}s.
\end{align*}
where $t\in[t_{k}^{i},t_{k+1}^{i})$.

For any neuron $v_{i}~(i=\oneton)$, if there are no events in $[t^{i}_{k}, t^{i}_{k}+T)$, the compulsory criterion will be triggered, that is, $t^{i}_{k+1}=t^{i}_{k}+T$. Otherwise, if there is a triggering event in $[t^{i}_{k},t^{i}_{k}+T)$, according to the autonomy criterion, it satisfies $t_{k}^{i}\leqslant t_{k+1}^{i}<t^{i}_{k}+T$. That is to say, it always satisfies $t_{k+1}^{i}-t^{i}_{k}\leqslant T$. Noting $
\frac{d}{dt}F_{i}(x_{t})=-d_{i}F_{i}(x_{t})$
if there is no event occurring at $t$.
Then, there exists some $\sigma>0$ such that
\begin{align*}
0<\frac
{\sum\limits_{i=1}^{n}\Big|F_{i}\big(x_{t_{k}^{i}}\big)\Big|^{2}}
{\sum\limits_{i=1}^{n}\Big|F_{i}\big(x_{t_{k+1}^{i}}\big)\Big|^{2}}
\leqslant
\frac
{\sum\limits_{i=1}^{n}\Big|F_{i}\big(x_{t_{k}^{i}}\big)\Big|^{2}}
{\sum\limits_{i=1}^{n}\Big|F_{i}\big(x_{t_{k}^{i}+T}\big)\Big|^{2}}
\leqslant\sigma
\end{align*}
for all $i=\oneton$ and $k=\zerotoinfty$, with taking $\sigma={\rm e}^{2d_{\max}T}$. Moreover, since $|F_{i}(x(t))|$ is decreasing on $[t_{k}^{i},t_{k+1}^{i})$ if there are no events occuring during this period, we obtain
\begin{align*}
\frac{\sum\limits_{i=1}^{n}\Big|F_{i}\big(x_{t_{k+1}^{i}}\big)\Big|^{2}}{n}
\leqslant\delta(t)
\leqslant\frac{\sum\limits_{i=1}^{n}\Big|F_{i}\big(x_{t_{k}^{i}}\big)\Big|^{2}}{\sum\limits_{i=1}^{n}{\rm e}^{-2d_{\max}T}}
\end{align*}
where $d_{\max}=\max_{i=\oneton}\{d_{i}\}$, which implies
\begin{align*}
\frac{\delta(t)}{\delta(t_{k+1}^{i})}
\leqslant
\frac{\sum\limits_{i=1}^{n}\Big|F_{i}\big(x_{t_{k}^{i}}\big)\Big|^{2}}{\sum\limits_{i=1}^{n}{\rm e}^{-2d_{i}T}}
\frac{n}{\sum\limits_{i=1}^{n}\Big|F_{i}\big(x_{t_{k+1}^{i}}\big)\Big|^{2}}
\leqslant\sigma{\rm e}^{2d_{\max}T}
\end{align*}
for any $t\in[t_{k}^{i},t_{k+1}^{i})$. Based on the autonomy criterion of the updating rule in Theorem \ref{PrimaryRule}, the event will not trigger until $|e_{i}(t)|=\gamma\Psi_{i}(t)$ at time point $t=t_{k+1}^{i}>t_{k}^{i}$. Hence, for each $x(0)\in\R^{n}\backslash\mathcal O$, it holds
\begin{align*}
\gamma\,{\rm e}^{-d_{i}(t_{k+1}^{i}-t_{k}^{i})}
&=\frac{\big|e_{i}(t_{k+1}^{i})\big|}{\sqrt{\delta(t_{k+1}^{i})}}\leqslant
 \mathcal M\int_{t_{k}^{i}}^{t_{k+1}^{i}}\frac{\sqrt{\delta(s)}}{\sqrt{\delta(t_{k+1}^{i})}}{\rm d}s\\
&\leqslant\mathcal M\sqrt{\sigma}\,{\rm e}^{d_{\max}T}(t_{k+1}^{i}-t_{k}^{i})
\end{align*}
Noting that the equation
\begin{align*}
\frac{\gamma\,e^{-d_{i}\eta_{i}}}{\mathcal M\sqrt{\sigma}\,{\rm e}^{d_{\max}T}}=\eta_{i}
\end{align*}
with $\eta_{i}=t_{k+1}^{i}-t_{k}^{i}$ possesses a positive solution of $\eta_{i}$, we can assert that for all  neurons $v_{i}~(i=\oneton)$, the next inter-event time has a common positive lower-bound which follows
\begin{align}\label{LowBound}
\eta=\min_{i\in\{\oneton\}}\left\{\eta_{i}:
\frac{\gamma\,{\rm e}^{-d_{i}\eta_{i}}}{\mathcal M\sqrt{\sigma}\,{\rm e}^{d_{\max}T}}=\eta_{i}
\right\}.
\end{align}
This implies that $|e_{i}(t)|=\gamma\Psi_{i}(t)$ holds at some $t=t_{k+1}^{i}$, which must satisfy $t_{k+1}^{i}-t_{k+1}^{i}\geqslant\eta$ for all $i=\oneton$ and $k=\zerotoinfty$.

Since $\eta$ and $T$ are uniform for all the neurons, the next triggering time point $t_{k+1}^{i}$ satisfies $t_{k+1}^{i}\geqslant t_{k}^{i}+\min\{\eta,T\}$ for all $i=\oneton$ and $k=\zerotoinfty$. Hence, the next inter-event interval of each neuron is lower bounded by a common positive constant, which means the absence of the {\it alternate triggering situation} in the Zeno behavior (P\ref{P2}) is proved.
\end{IEEEproof}
To sum up, we have excluded both the {\it continuous triggering situation} and {\it alternate triggering situation} in the Zeno behavior, when the event-triggered rule is taken into account. Therefore, we can claim that there is no Zeno behavior for all the neurons.

After proving exclusion of Zeno behaviors, we are at the stage to conclude $\lim_{k\to\infty}t^{i}_{k}=\infty$ for all $i=\oneton$. This implies the following summary result.
\begin{theorem}\label{Main}
Under the event-trigger rule described in Theorem \ref{PrimaryRule}, system (\ref{mg}) is almost sure stable.
\end{theorem}
\begin{IEEEproof}
In fact, from Theorems \ref{MSTheorem} and \ref{Zeno}, one can conclude that for all initial values except a set of zero measure, the trajectory of system (\ref{mg}) possesses discontinuous the trigger events with inter-event interval a positive low bounded. This implies that the solution of Cauchy problem of system (\ref{mg}) with these initial values exists for the duration $[0,\infty)$ and $\lim_{k\to\infty}t^{i}_{k}=+\infty$ for all $i=\oneton$. From Theorem \ref{PrimaryRule}, it converges to certain equilibrium on $\mathcal S$. Therefore, we have proved this theorem.
\end{IEEEproof}

\section{Discrete-time monitoring}\label{Monitoring}
The continuous monitoring strategy for Theorem \ref{PrimaryRule} may be costly since the states of the neurons should be observed simultaneously. An alternative method is to predict the triggering time point when inequality \eqref{PrimaryRule1} does not hold and update the triggering time accordingly.

For any neuron $v_{i}~(i=\oneton)$, according to the current event timing $t_{k}^{i}$, its state can be formulated as
\begin{flalign}\label{StateFormula}\raisetag{40pt}
\begin{cases}
x_{i}(t)=x_{i}(t_{k}^{\bm*})+\dfrac{1}{d_{i}}
         \bigg\{d_{i}x_{i}(t_{k}^{\bm*})+\Big[\nabla f\big(y(t_{k}^{i})\big)\Big]_{i}-\theta_{i}\bigg\}\\[10pt]
\hspace{7.5ex}\times\Big[{\rm e}^{-d_{i}(t-t_{k}^{\bm*})}-1\Big]\\[7pt]
y_{i}(t)=g_{i}\big(\lambda_ix_i(t)\big)\\[3pt]
\end{cases}&&
\end{flalign}
for $t_{k}^{\bm*}<t<t_{k+1}^{i}$, where $t_{k}^{\bm*}=\max_{j}\big\{t_{k}^{j}\big\}$ is the newest timing of all other neurons and $t_{k+1}^{i}$ is the next triggering time point at which neuron $v_{i}$ happens the triggering event. Then, solving the following maximization problem
\begin{align}
\label{Maximization}
\Delta t_{k}^{i}
=\max_{t\in(t_{k}^{\bm*},t_{k+1}^{i})}\Big\{t-t_{k}^{\bm*}:\big|e_{i}(t)\big|\leqslant\gamma\Psi_{i}(t)\Big\},
\end{align}
we have the following prediction algorithm (Algorithm \ref{Algorithm}) for the next triggering time point.

With the information of each neuron at time $t^{i}_{k}$ and the proper parameters $\gamma$, search the observation time $\Delta t_{k}^{i}$ by \eqref{Maximization} at first. If no triggering events occur in all neurons during $(t_{k}^{i},t_{k}^{\bm*}+\Delta t_{k}^{i})$, the neuron $v_{i}$ triggers at time $t_{k}^{\bm*}+\Delta t_{k}^{i}$ and record as the next triggering event time $t_{k+1}^{i}$, that is $t_{k+1}^{i}=t_{k}^{\bm*}+\Delta t_{k}^{i}$. Renew the neuron $v_{i}$'s state and send the renewed information to the other neurons. The prediction of neuron $v_{i}$ is finished. If some other neuron triggers at time $t\in(t_{k}^{i},t_{k}^{\bm*}+\Delta t_{k}^{i})$, update $t_{k}^{\bm*}$ in state formula \eqref{StateFormula} and go back to find a new observation time $t_{k}^{\bm*}+\Delta t_{k}^{i}$ by solving the maximization problem \eqref{Maximization}.

\begin{algorithm}[ht]
\caption{Prediction for the next triggering time point $t_{k+1}^{i}$}
\label{Algorithm}
\hspace*{\fill}
\begin{algorithmic}[1]
\Require
\State Initialize $\gamma>0$, $t_{k}^{\bm*}\gets \max_{j}t_{k}^{j}$
\State Input $x_{i}(t)\gets x_{i}(t_{k}^{i})$ for all $i=\oneton$
\Ensure
\State $\text{Flag}\gets 0$
\While {Flag $=0$} 
\State Search $\Delta t_{k}^{i}$ by the strategy \eqref{Maximization}
\State $\tau\gets t_{k}^{\bm*}+\Delta t_{k}^{i}$
\If{No neurons  trigger during $(t_{k}^{i},\tau)$}
\State $v_{i}$ triggers at time $t_{k+1}^{i}=\tau$
\State $v_{i}$ renew its state information $x_{i}(t_{k+1}^{i})$
\State $v_{i}$ sends the state information to the other neurons
\State $\text{Flag}\gets1$ %
\Else
\State Update $t_{k}^{\bm*}$ in the state formula \eqref{StateFormula}
\EndIf
\EndWhile
\State\Return{$t_{k+1}^{i}$}
\end{algorithmic}
\end{algorithm}

In addition, when neuron $v_{i}$ updates its observation time $\Delta t_{k}^{i}$, the triggering time predictions of the neurons will be affected. Therefore, besides the state formula \eqref{StateFormula} and the maximization problem \eqref{Maximization} as given before, each neuron should take their triggering event time whenever any of the other neurons renews and broadcasts its state information. In other word, if one neuron updates its triggering event time, it is mandatory to inform the other neurons.

\begin{remark}
This monitoring scheme via the state formula \eqref{StateFormula} may lose the high-level efficiency of the convergence, because it abandons the continuous adjustment on $\delta(t)$ in Eq. \eqref{normalization}. But the advantage is that a discrete-time inspection on $x(t)$ can be introduced to ensure the convergence in Theorem \ref{PrimaryRule}, which can reduce the monitoring times and costs.
\end{remark}

\section{Examples}\label{sec5}
In this section, two numerical examples are given to demonstrate the effectiveness of the presented results and the application. Example 1 is a 5-dimension system which illustrates our theoretical results and Example 2 is a 3-dimension system which compares the continuous monitoring with the discrete-time monitoring.

\noindent{\bf Example 1:} Consider a 5-dimension analytic neural network with
\begin{align}
\label{Example1}
f(y)=\sum_{i=1}^{5}\left(\frac{3}{4}y_{i}^4-y_{i}^3\right)-\frac{1}{2}y^{\top}Wy+\sum_{i=1}^{5}y_{i}
\end{align}
where $D=\Lambda=\text{diag}\{1,1,1,1,1\}$, $\theta=[1,-1,1,-1,1]^{\top}$ and
\begin{align*}
W=\setlength{\arraycolsep}{5pt}
\begin{bmatrix}
~~3.919 & ~3.948 & ~~2.564 & ~~3.204 & ~~0.156\\
 -4.672 & ~6.491 &  -4.117 &  -1.371 &  -0.501\\
~~4.011 & ~1.370 & ~~5.727 & ~~5.411 & ~~1.185\\
 -1.983 & ~1.656 &  -8.428 & ~~7.652 &  -7.694\\
~~1.282 & ~2.135 & ~~5.559 & ~~0.659 & ~~9.569
\end{bmatrix}.
\end{align*}

By the rule \eqref{PrimaryRule1}, Fig. \ref{Example1:Fig1} illustrates the state $x(t)$ converges to $x^{\bm*}=[-1.314,0.861,-1.709,0.580,-0.944]^{\top}$. The initial value $x(0)=[0.728,-0.769,1.770,-1.827,0.315]^{\top}$ is randomly selected in the domain $[-2,2]^{5}$ and $\gamma=0.3$. The triggering time points of each neuron are shown in Fig. \ref{Example1:Fig2}.

Take the different values of the parameter $\gamma$ from $0.1$ to $0.5$ by step $0.05$ and $\sqrt{\alpha}/\sqrt{\beta}=0.527$. The simulation results under the event-triggered rule are shown in Table \ref{Example1:Table}. $\eta$ is the theoretical lower-bound for the inter-event time of all the neurons calculated by \eqref{LowBound}. $\eta_{\,\rm sim}=\min_{i\in\{\oneton\}}\min_{k\in\{\zerotoinfty\}}(t_{k+1}^{i}-t_{k}^{i})$ is the actual value of the minimal length of inter-event time by simulation. $N$ is the average number of triggering events over the neurons and $T_{\,\rm first}$ stands for the first time when $\|x(t)-x^{\bm*}\|\leqslant 0.001$. All the results in the table are average over $50$ independent simulations.

It can be seen that the actual minimal inter-event time $\eta_{\,\rm sim}$ is always larger than the corresponding theoretical lower-bound $\eta$. This implies that we have excluded the Zeno behavior with the lower-bound $\eta$ for all the neurons. Moreover, The average number of triggering events $N$ decreases while the first convergent time $T_{\,\rm first}$ increases with $\gamma$ increasing from $0.1$ to $0.5$ by step $0.05$.\\

\begin{figure}[H]
\centering
\subfigure[Dynamics of components in system \eqref{Example1} under the updating rule \eqref{PrimaryRule1}.]
{\label{Example1:Fig1}\includegraphics[width=0.49\textwidth]{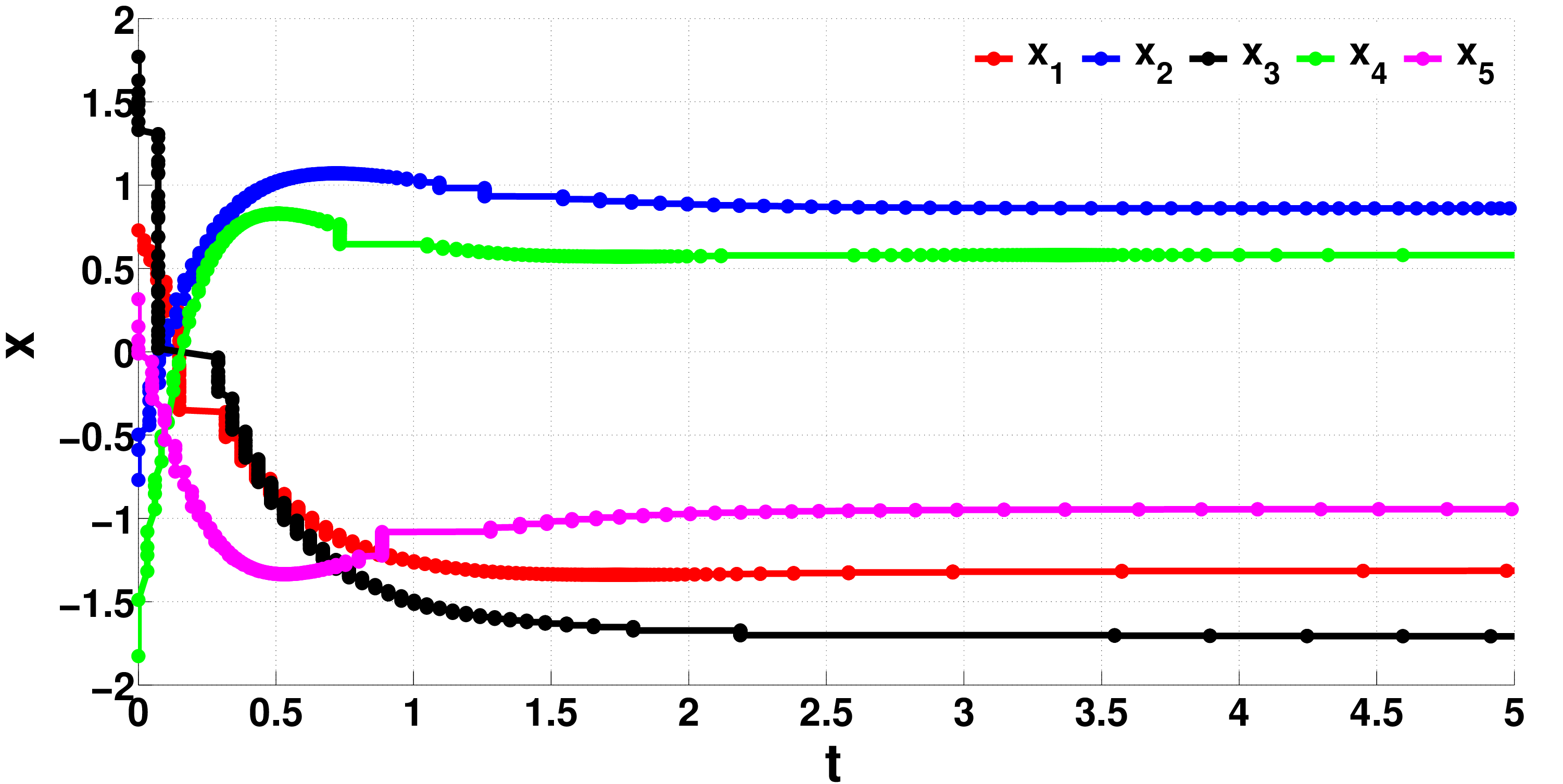}}\\
\subfigure[Time slots of triggering events at each neuron for stability.]
{\label{Example1:Fig2}\includegraphics[width=0.49\textwidth]{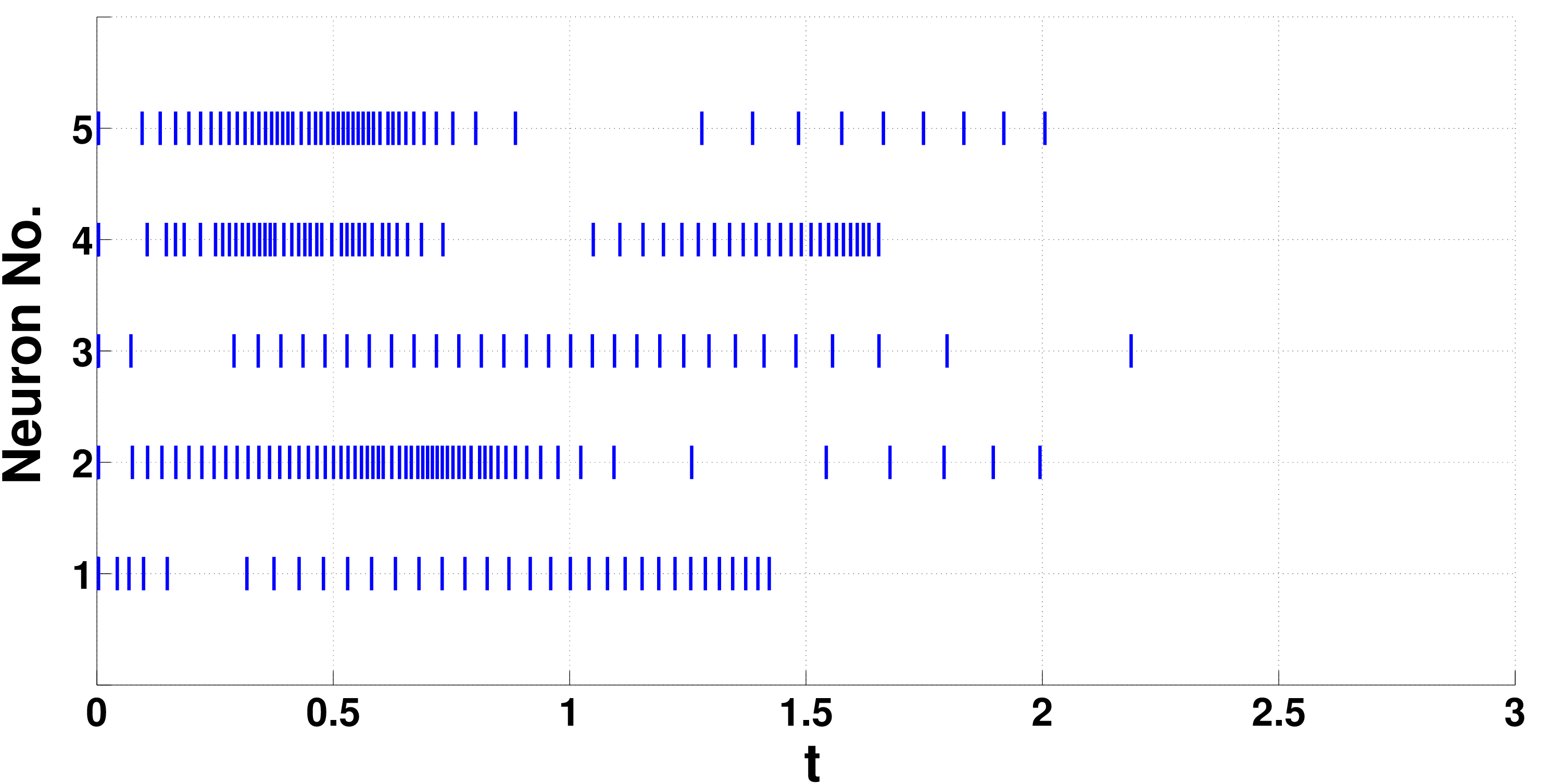}}\\
\caption{Dynamics of the system \eqref{Example1} and the triggering time points $\{t_{k}^{i}\}_{k=0}^{+\infty}$.}
\label{Example11}
\end{figure}

\begin{table}[H]
\centering
\renewcommand\arraystretch{1.3}\setlength{\tabcolsep}{10pt}\small
\caption{Simulation results under the event-triggered rule with $d_{1}=d_{2}=1$, $T=0.03$, $\sigma={\rm e}^{2d_{\max}T}$ and $M=\sqrt{5}/{4}$}
\begin{tabular}{c|c|c|c|c}
\hline\rule{0pt}{2ex}
$\gamma$ & $\eta_{\,\rm sim}$ & $\eta$  & $N$ & $T_{\,\rm first}$ \\\hline
  0.10   & 0.00490 & 0.00443  &82.9   &1.948  \\\hline
  0.15   & 0.00712 & 0.00665  &68.6   &1.981  \\\hline
  0.20   & 0.00908 & 0.00887  &59.5   &2.091  \\\hline
  0.25   & 0.01476 & 0.01108  &52.9   &2.152  \\\hline
  0.30   & 0.01505 & 0.01330  &47.6   &2.188  \\\hline
  0.35   & 0.01789 & 0.01552  &40.5   &2.203  \\\hline
  0.40   & 0.01898 & 0.01774  &35.9   &2.362  \\\hline
  0.45   & 0.02124 & 0.01995  &32.6   &2.525  \\\hline
  0.50   & 0.02351 & 0.02217  &28.7   &2.718  \\\hline
\end{tabular}\label{Example1:Table}
\end{table}

\ignore{
\begin{figure}[H]
\centerline{
\includegraphics[width=0.5\textwidth]{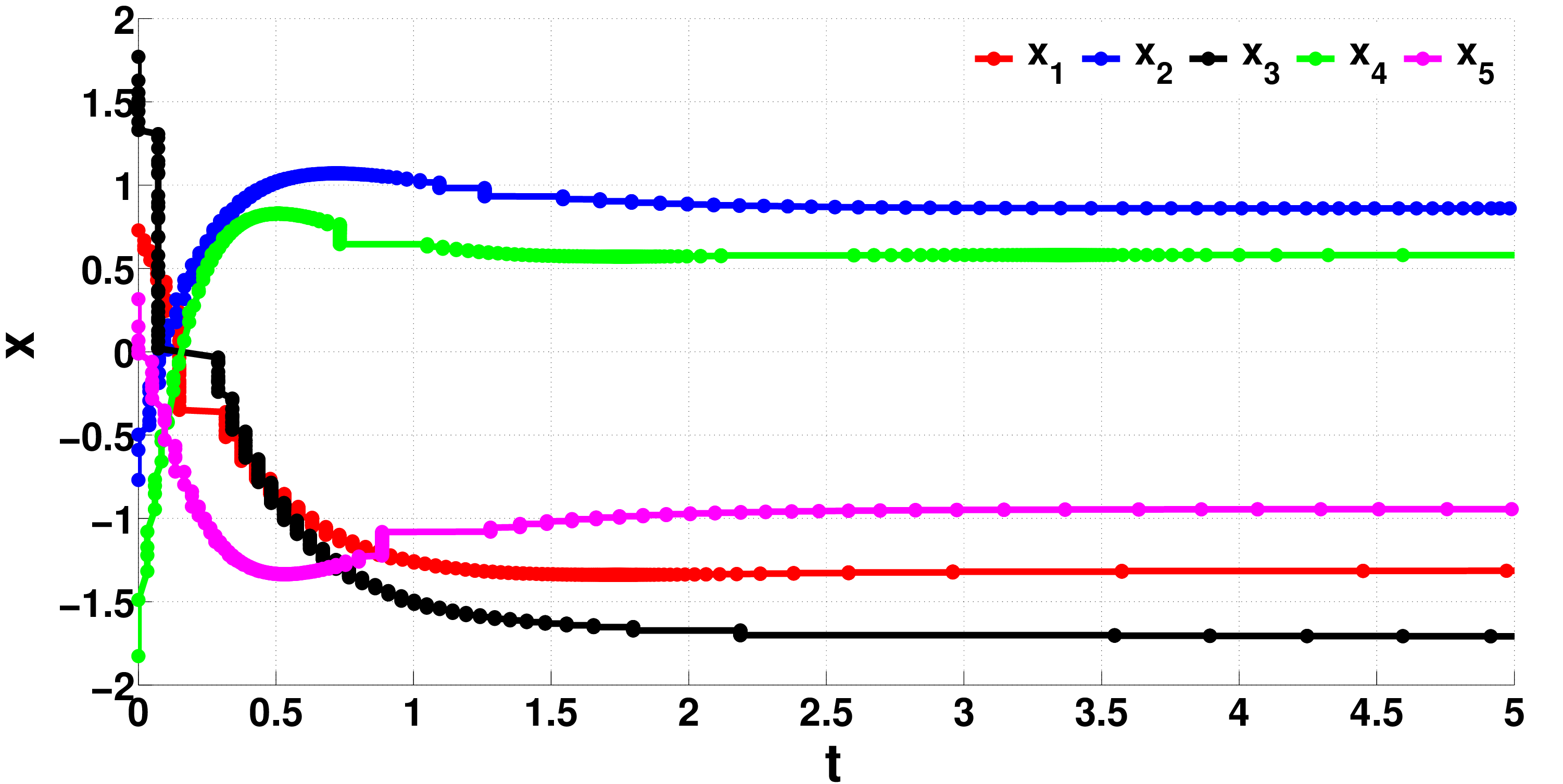}}
\caption{Dynamics of components in system \eqref{Example1} under the event-triggered rule \eqref{PrimaryRule1}.}
\label{Example1:Fig1}
\end{figure}
\begin{figure}[H]
\centerline{
\includegraphics[width=0.5\textwidth]{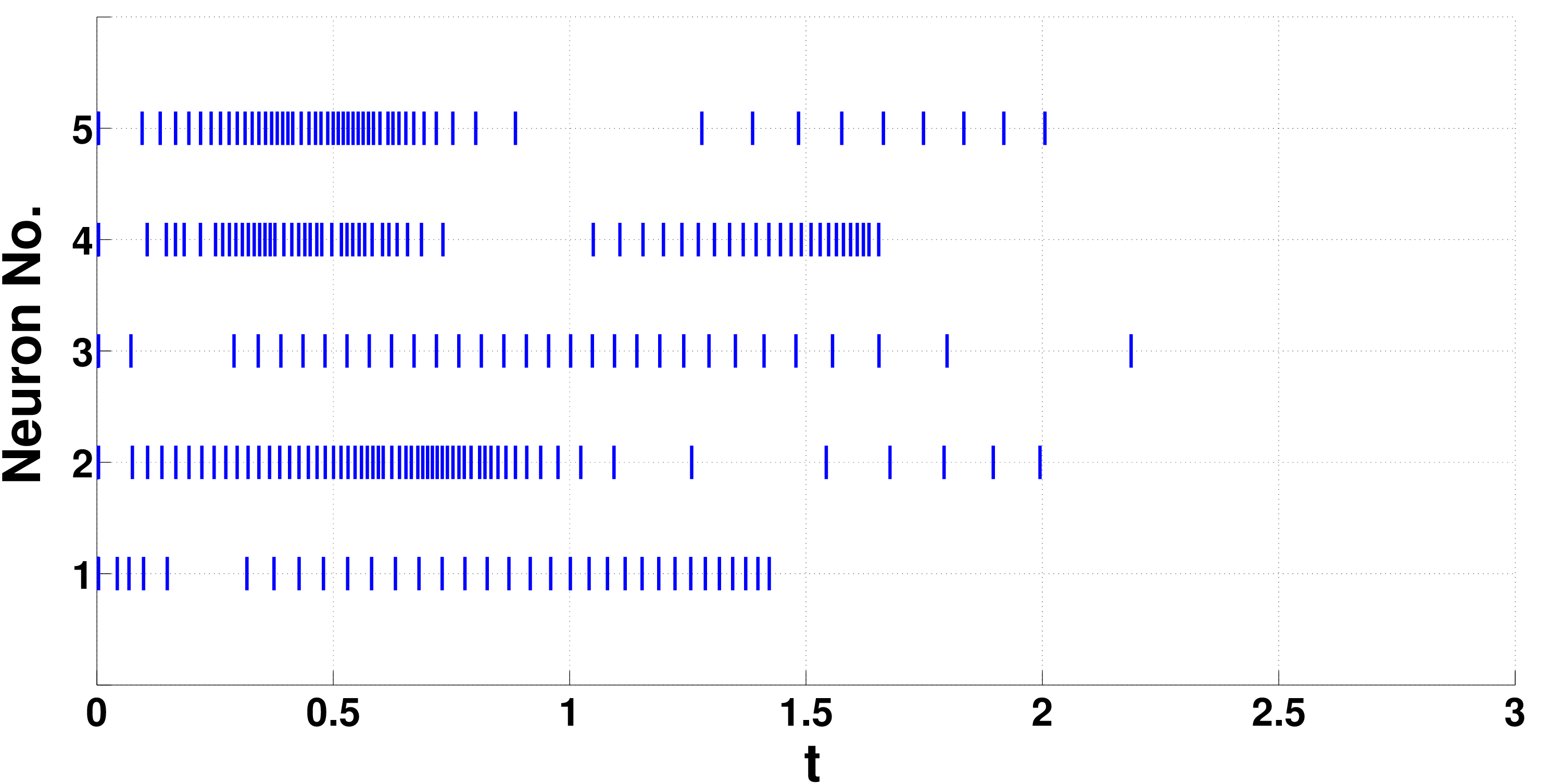}}
\caption{Time slots of triggering events at each neuron for stability.}
\label{Example1:Fig2}
\end{figure}}

\noindent{\bf Example 2:}
Consider a 3-dimension neural network \eqref{mg} with
\begin{align}\label{Example2}
f(y)=\sum_{i=1}^{3}\bigg(\frac{1}{2}y_{i}^{4}-y_{i}^{3}\bigg)-\frac{1}{2}y^{\top}Wy+\sum_{i=1}^{3}y_{i},
\end{align}
where $D=\Lambda=\text{diag}\{1,1,1,1,1\}$, $\theta=[1,-1,1,-1,1]^{\top}$ and
\begin{align*}
W= \setlength{\arraycolsep}{5pt}\begin{bmatrix} 3 & 2.5 & 2\\ 2 & 2 & 3\\ 3 & 2 & 2.5 \end{bmatrix}
\end{align*}

\begin{figure}[H]
\centering
\subfigure[Dynamics of components in system \eqref{Example2} by continuous-time monitoring under the event triggering rule \eqref{PrimaryRule1}.]
{\label{ContinuousDynamics}\includegraphics[width=0.49\textwidth]{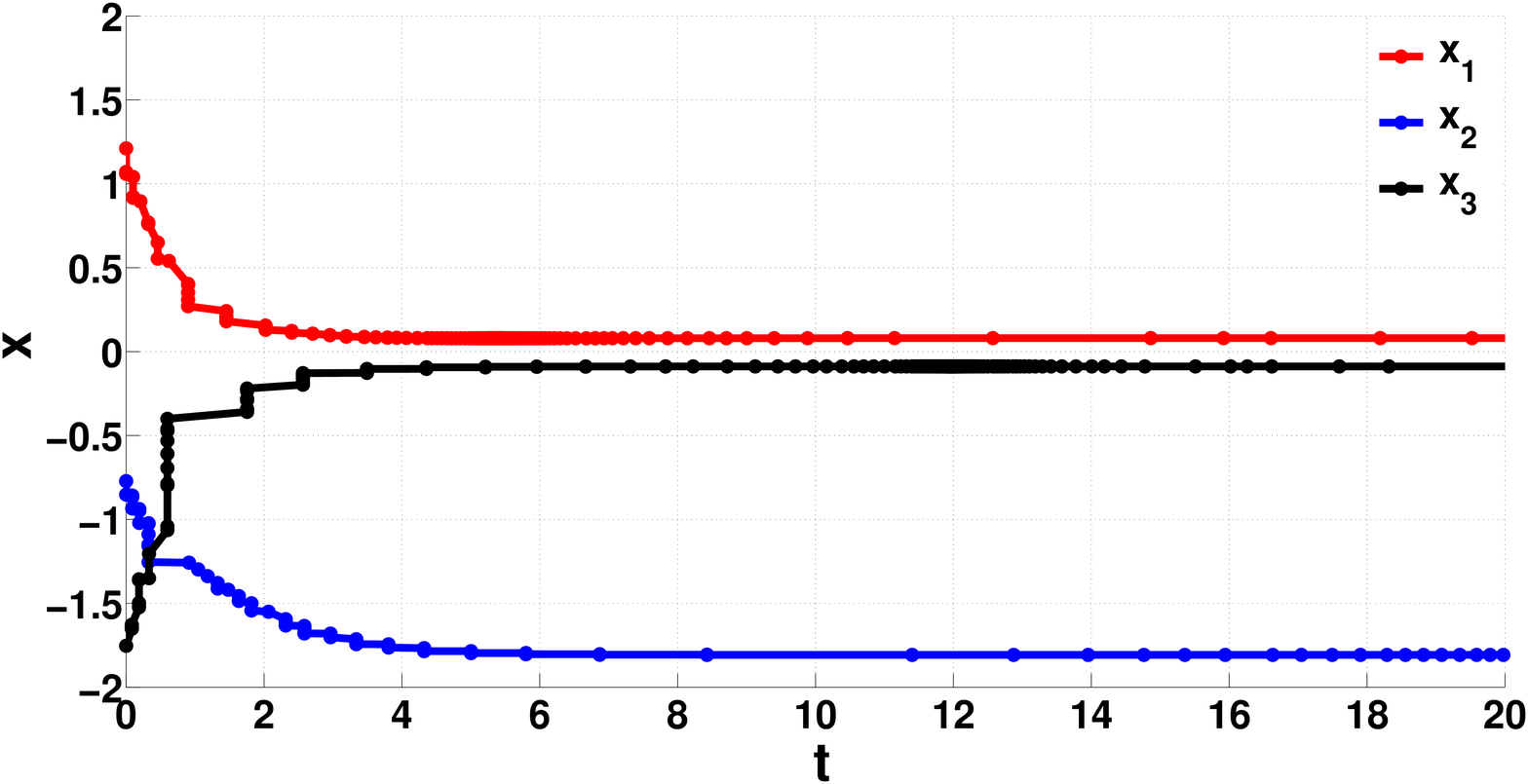}}\\
\subfigure[Time slots of triggering events by continuous-time monitoring.]
{\label{ContinuousTime}\includegraphics[width=0.49\textwidth]{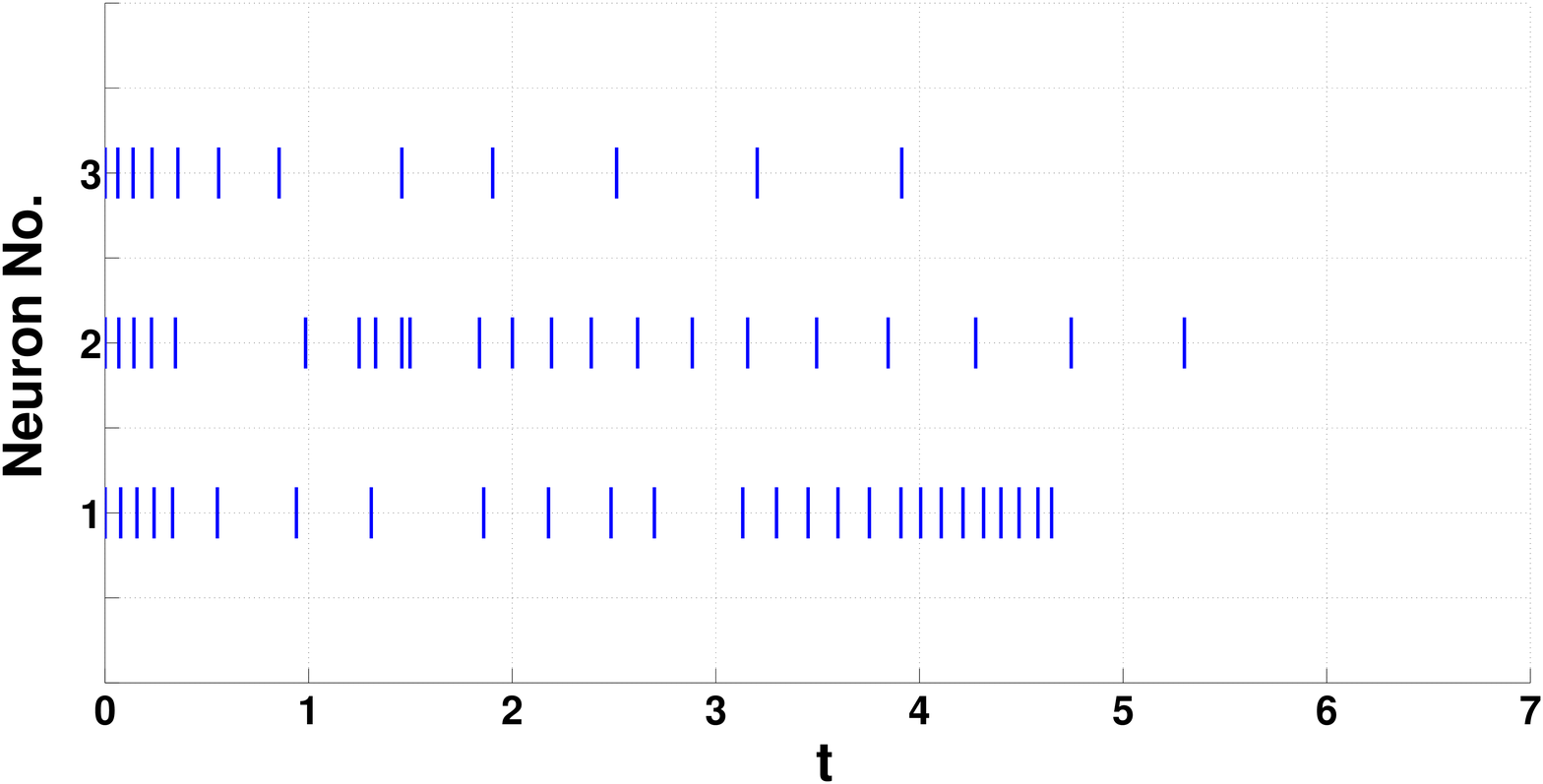}}\\
\subfigure[Dynamics of components in system \eqref{Example2} by discrete-time monitoring under the event triggering rule \eqref{PrimaryRule1}.]
{\label{DiscreteDynamics}\includegraphics[width=0.49\textwidth]{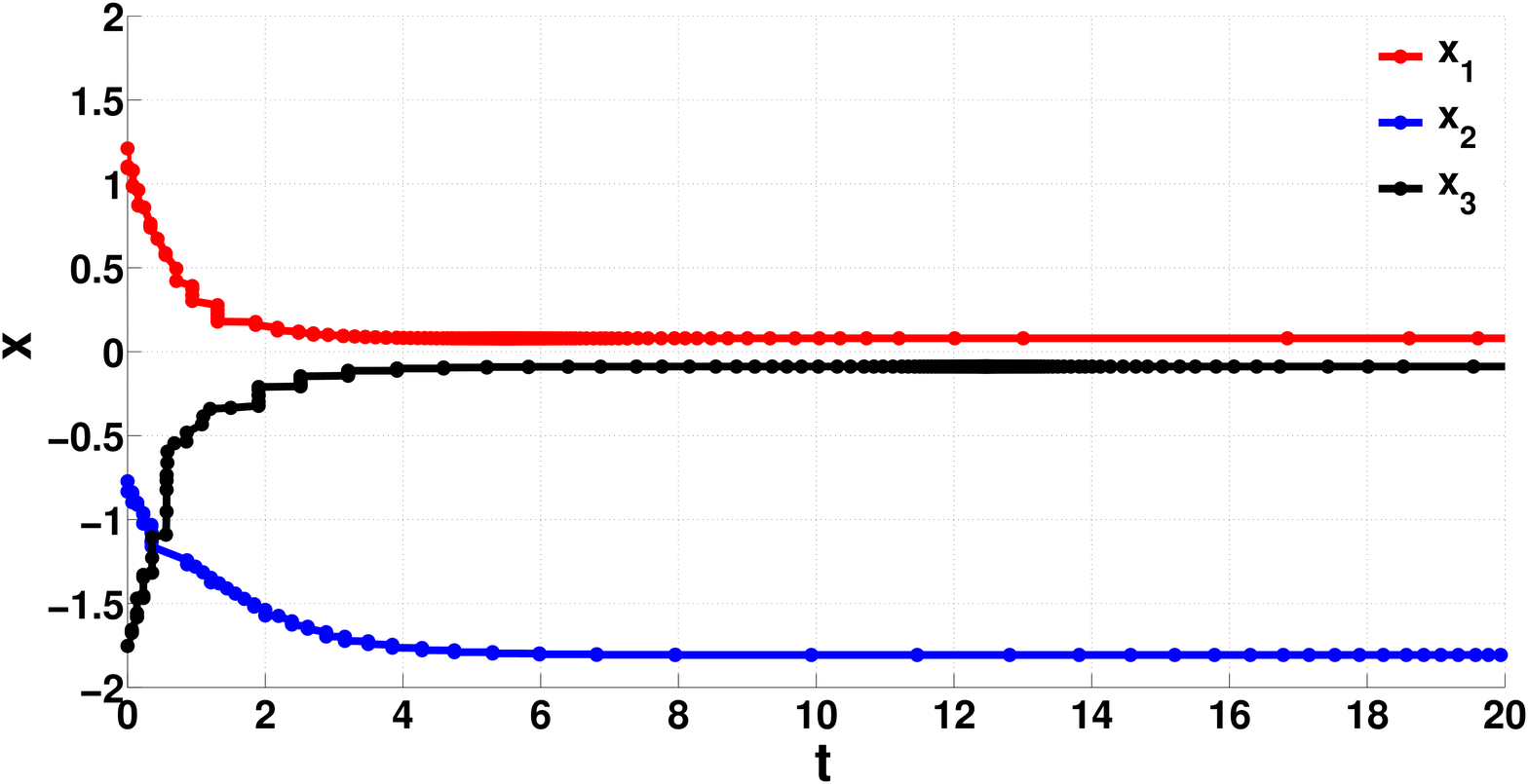}}\\
\subfigure[Time slots of triggering events by discrete-time monitoring.]
{\label{DiscreteTime}\includegraphics[width=0.49\textwidth]{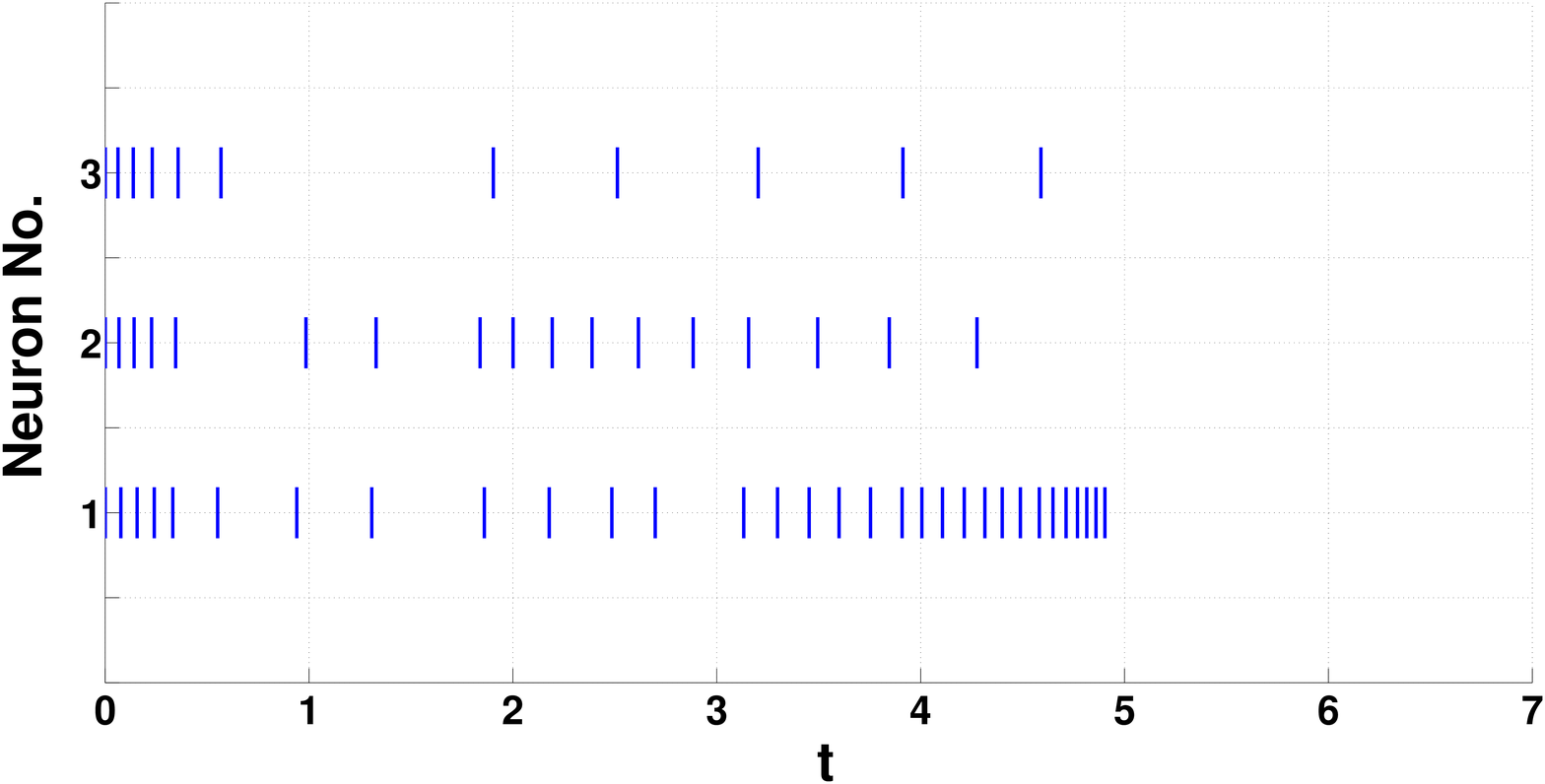}}
\caption{Dynamics of the system \eqref{Example2} and the triggering time points $\{t_{k}^{i}\}_{k=0}^{+\infty}$.}
\label{Error}
\end{figure}

According to the event-triggered rule \eqref{PrimaryRule1} in Theorem \ref{PrimaryRule}, Fig. \ref{ContinuousDynamics} shows that the state $x(t)$ converges to the equilibrium by continuous monitoring and Fig. \ref{DiscreteDynamics} indicates the convergence by discrete-time monitoring.
With $x(0)=[1.211,-0.772,-1.7530]^{\top}$ and $\gamma=0.5$, the equilibrium is $x^{\bm*}=[0.080,-1.807,-0.088]^{\top}$.  The time slots of the triggered events of each neuron by continuous-time and discrete-time monitoring are illustrated in Figs. \ref{ContinuousTime} and \ref{DiscreteTime}.
We record $\eta_{\,\rm sim}$, $\eta$, $N$ and $T_{\,\rm first}$ with different values of $\gamma$ by both continuous-time and discrete-time monitoring. The results shown in Table \ref{Example2:Table1} are average $50$ independent simulations. It can be seen that $\eta_{\,\rm sim}$ is larger than the theoretical lower-bound $\eta$ by both continuous-time and discrete-time monitoring, which implies that the Zeno behavior is excluded for all the neurons. $N$ and $T_{\,\rm first}$ from two monitoring are similar to those in Example 1. $N$ decreases and $T_{\,\rm first}$ increases with $\gamma$ increasing from $0.1$ to $0.5$ by step $0.05$ and
$\sqrt{\alpha}/\sqrt{\beta}=0.527$.
\begin{table}[H]
\centering
\renewcommand\arraystretch{1.3}\setlength{\tabcolsep}{10pt}\small
\caption{Simulation results under the event-triggered rule by continuous monitoring and discrete-time monitoring with
$d_{1}=d_{2}=1$, $T=3$, $\sigma={\rm e}^{2d_{\max}T}$ and $M=\sqrt{26}/80$}
\begin{tabular}{c|c|c|c|c}
\hline\rule{0pt}{2.0ex}
\multirow{2}*{$\gamma$} &\multicolumn{3}{c|}{Continuous-time} & \multirow{2}*{$\eta$}\\\cline{2-4}
                        & $\eta_{\,\rm sim}$ & $N$  & $T_{\,\rm first}$   &   \\\hline
  0.10   & 0.01320  & 38.3 & 4.356  & 0.00423    \\\hline
  0.20   & 0.01760  & 36.7 & 4.376  & 0.00846     \\\hline
  0.30   & 0.01978  & 29.4 & 4.532  & 0.01268     \\\hline
  0.40   & 0.02756  & 25.8 & 4.608  & 0.01691     \\\hline
  0.50   & 0.04548  & 20.3 & 4.656  & 0.02114     \\\hline
\multirow{2}*{$\gamma$} &\multicolumn{3}{c|}{Discrete-time} & \multirow{2}*{$\eta$}\\\cline{2-4}
                        & $\eta_{\,\rm sim}$ & $N$  & $T_{\,\rm first}$   &   \\\hline
  0.10   & 0.01320  & 37.2 & 4.254  & 0.00423    \\\hline
  0.20   & 0.01760  & 35.7 & 4.344  & 0.00846     \\\hline
  0.30   & 0.01978  & 28.1 & 4.513  & 0.01268     \\\hline
  0.40   & 0.02756  & 23.7 & 4.598  & 0.01691     \\\hline
  0.50   & 0.04548  & 19.3 & 4.652  & 0.02114     \\\hline
\end{tabular}\label{Example2:Table1}
\end{table}

\begin{figure}[ht]
\centerline{
\includegraphics[width=0.5\textwidth]{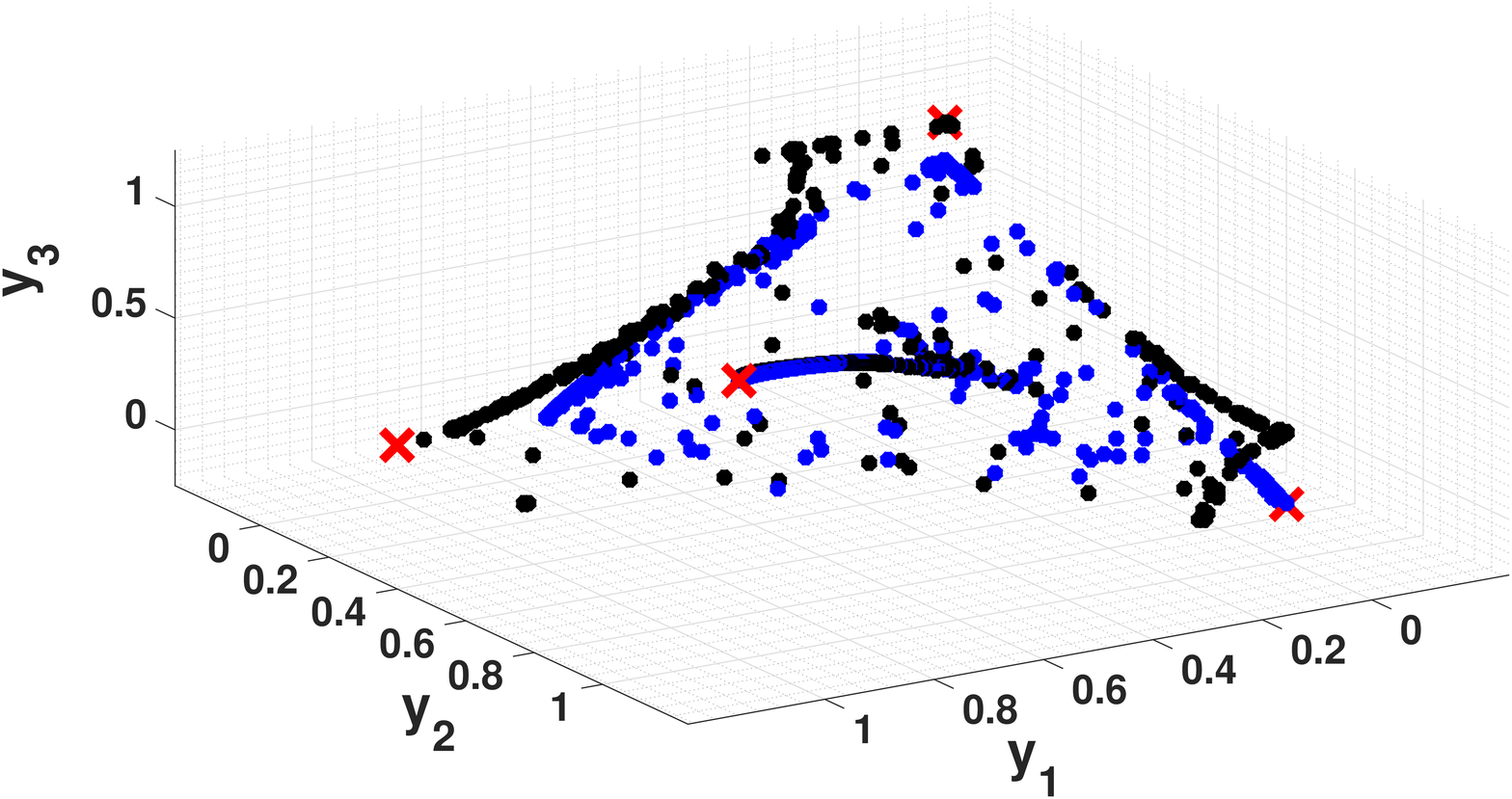}}
\caption{The limit $\overline{y}(\Lambda)$ converges to two local minimum points $[1,0,0]^{\top}$ and $[0,1,0]^{\top}$ with a random input $\theta$ with $|\theta|<0.001$, $\gamma=0.5$, and random initial data in the domain $[-5,5]^{3}$. $\lambda_1=\lambda_2$ are picked from 0.01 to 100.}
\label{Example2:Fig3}
\end{figure}

According to the definition of Lyapunov (or energy) function \eqref{Ly}, if the input $\|\theta\|$ takes a sufficient small value and $\lambda_i\rightarrow+\infty$ for $i=1,2,3$, then $L(x)\approx f(y)$.
Thus, as an application, the system \eqref{Example2} with the event-triggered rule  \eqref{PrimaryRule1} in Theorem \ref{PrimaryRule} can be used to seek the local minimum point of $f(y)$ over $\{0,1\}^{3}$. Denote
\begin{align*}
\overline{y}(\Lambda)=\lim_{t\rightarrow+\infty}g\big(\Lambda x(t)\big),
\end{align*}
where $x(t)$ is the trajectory of the system \eqref{Example2}. $\overline{y}(\Lambda)$ is the local minimum point of $H(y)$ where
\begin{align*}
H(y)=\sum_{i=1}^{3}\bigg(\frac{1}{2}y_{i}^{4}-y_{i}^{3}\bigg)-\frac{1}{2}y^{\top}Wy+\sum_{i=1}^{3}y_{i},
\end{align*}
Fig. \ref{Example2:Fig3} shows that the limit $\overline{y}(\Lambda)$ converges to local minimum points $[1,0,0]^{\top}$ and $[0,1,0]^{\top}$ when $\lambda_{i}\to+\infty~(i=1,2,3)$.  The initial value $x(0)$ for each simulation is chosen randomly in the domain $[-5,5]^{3}$ and $\gamma=0.5$.

\section{Conclusion}\label{sec6}
In this paper, the event-triggering rule for discrete-time synaptic feedbacks in a class of analytic neural network was proposed and proved to guarantee neural networks to be almost sure stable. In addition, the Zeno behaviors can be prove to be excluded. By these asynchronous event-triggering rules, the synaptic information exchanging frequency between neurons are significantly reduced. The main technique of proving almost stability is finite-length of trajectory and the {\L}ojasiewicz inequality \cite{Mfa}. Two numerical examples have been provided to demonstrate the effectiveness of the theoretical results. It has also been shown by these examples, following the routine in \cite{Wll} and the proposed updating rule can reduce the cost of synaptic interactions between neurons. One step further, our future work will include the self-triggered formulation and event-triggered stability of other more general systems as well as their application in dynamic optimisation.


\begin{thebibliography}{22}

\bibitem{Aapt}
A. Anta and P. Tabuada,
``Self-triggered stabilization of homogeneous control systems,''
in {\it Proceedings of American Control Conference}, Seattle, WA, Jun. 2008, pp. 4129-4134.

\bibitem{Aap}
A. Anta and P. Tabuada,
``To sample or not to sample: self-triggered control for nonlinear systems,''
{\it IEEE Trans. Autom. Control}, vol. 55, pp. 2030-2042, Sep. 2010.

\bibitem{AmSh}
A. Molin and S. Hirche,
``Suboptimal Event-Based Control of Linear Systems Over Lossy Channels Estimation and Control of Networked Systems,''
in {\it Proceedings of the 2nd IFAC Workshop on Distributed Estimation and Control in Networked Systems}, Annecy, France, Sep. 2010, pp. 5560.

\bibitem{Apm}
A. P. Morse,
``The behaviour of a function on its critical set,''
{\it Ann. Math.}, vol. 40, no. 1, pp. 62-70, 1939.

\bibitem{As}
A. Sard,
``The measure of the critical values of differentiable maps,''
{\it Bull. Amer. Math. Soc.}, vol. 48, no. 12, pp. 883-890, 1942.

\bibitem{Dvd}
D. V. Dimarogonas, E. Frazzoli, and K. H. Johansson,
``Distributed event-triggered control for multi-agent systems,''
{\it IEEE Trans. Autom. Control}, vol. 57, pp. 1291-1297, May. 2012.

\bibitem{Ega}
E. Garcia and P. J. Antsaklis,
``Model-based event-triggered control with time-varying network delays,''
in {\it Proceedings of the 50th IEEE Conference on Decision and Control and European Control Conference}, Orlando, Fl, Dec. 2011, pp. 1650-1655.

\bibitem{Gss}
G. S. Seyboth, D. V. Dimarogonas, and K. H. Johansson,
``Event-based broadcasting for multi-agent average consensus,''
{\it Automatica}, vol. 49, no. 1, pp. 245-252, 2013.

\bibitem{Jdc}
J. D. Cao and J. Wang,
``Global asymptotic stability of a general class of recurrent neural networks with time-varying delays,''
{\it IEEE Trans. Circuit Syst. I, Fundam. Theory Appl.}, vol. 50, pp. 34-44, Jan. 2003.

\bibitem{Jjh1}
J. J. Hopfield,
``Neural networks and physical systems with emergent collective computational abilities,''
{\it Proc. Nat. Acad. Sci.}, vol. 79, no. 8, pp. 2554-2558, 1982.

\bibitem{Jjh}
J. J. Hopfield,
``Neurons with graded response have collective computa-tional properties like those of two-state neurons,''
{\it Proc. Nat. Acad. Sci.}, vol. 81, no. 10, pp. 3088-3092, 1984.

\bibitem{Jkh}
J. K. Hale,
{\it Ordinary Differential Equations.} New York, NY: John Wiley \& Sons, Inc., 1980.

\bibitem{Jma}
J. Ma$\acute{n}$dziuk,
``Solving the travelling salesman problem with a Hopfield-type neural network.''
{\it Demonstratio Math.}, vol. 29, no. 1, pp. 219-231, 1996.

\bibitem{Kgv}
K. G. Vamvoudakis,
``An Online Actor/Critic Algorithm for Event-Triggered Optimal Control of Continuous-Time Nonlinear Systems,''
in {\it Proceedings of American Control Conference}, Portland, OR, Jun. 2014, pp. 1-6.

\bibitem{Khj}
K. H. Johansson, M. Egerstedt, J. Lygeros, and S. S. Sastry,
``On the regularization of zeno hybrid automata,''
{\it Systems and Control Letters}, vol. 38, no. 3, pp. 141-150, 1999.

\bibitem{Loc}
L. O. Chua and L. Yang,
``Cellular neural networks: Theory,''
{\it IEEE Trans. Circuits Syst.}, vol. 35, pp. 1257-1272, Oct. 1988.

\bibitem{Loc1}
L. O. Chua and L. Yang,
``Cellular neural networks: Application,''
{\it IEEE Trans. Circuits Syst.}, vol. 35, pp. 1273-1290, Oct. 1988.

\bibitem{Mac}
M. A. Cohen and S. Grossberg,
``Absolute stability of global pattern formation and parallel memory storage by competitive neural networks,''
{\it IEEE Trans. Syst., Man, Cybern.}, vol. 13, pp. 815-821, Sep. 1983.

\bibitem{Mfa1}
M. Forti and A. Tesi,
``New conditions for global stability of neural networks with application to linear and quadratic programming problems,''
{\it IEEE Trans. Circuits Syst. I, Reg. Papers}, vol. 42, pp. 354-366, Jul. 1995.

\bibitem{Mfa}
M. Forti and A. Tesi,
``Absolute stability of analytic neural networks: An approach based on finite trajectory length,''
{\it IEEE Trans. Circuits Syst. I, Reg. Papers}, vol. 51, pp. 2460-2469, Dec. 2004.

\bibitem{Mfa2}
M. Forti and A. Tesi,
``The {\L}ojasiewicz exponent at equilibrium point of a standard CNN is 1/2,''
{\it Int. J. Bifurc. Chaos}, vol. 16, no. 8, pp. 2191-2205, 2006.

\bibitem{Mfp}
M. Forti, P. Nistri, and M. Quincampoix,
``Convergence of neural networks for programming problems via a nonsmooth {\L}ojasiewicz inequality,''
{\it IEEE Trans. Neural Netw.}, vol. 17, pp. 1471-1486, Nov. 2006.

\bibitem{Mhi}
M. Hirsh,
``Convergence activation dynamics in continuous time networks,''
{\it Neural Netw.}, vol. 2, no. 5, pp. 331-349, 1989.

\bibitem{Mjm}
M. J. Manuel, A. Anta, and P. Tabuada,
``An ISS self-triggered implementation of linear controllers,''
{\it Automatica}, vol. 46, no. 8, pp. 1310-1314, 2010.

\bibitem{Mmp}
M. J. Manuel and P. Tabuada,
``Decentralized event-triggered control over wireless sensor/actuator networks,''
{\it IEEE Trans. Autom. Control}, vol. 56, pp. 2456-2461, Oct. 2011.

\bibitem{Mvi}
M. Vidyasagar,
``Minimum-seeking properties of analog neural networks with multilinear objective functions,''
{\it IEEE Trans. Automat. Control}, vol. 40, pp. 1359-1375, Aug. 1995.

\bibitem{Pta}
P. Tabuada,
``Event-triggered real-time scheduling of stabilizing control tasks,''
{\it IEEE Trans. Autom. Control}, vol. 52, pp. 1680-1685, Sep. 2007.

\bibitem{Slo}
S. {\L}ojasiewicz,
``Une propriet$\acute{e}$ topologique des sous-ensembles analy-tiques r$\acute{e}$els,''
in {\it Colloques internationaux du C.N.R.S. les $\acute{e}$quations aux d$\acute{e}$rive$\acute{e}$s partielles}, Paris, France, 1963, pp. 87-89.

\bibitem{Slo1}
S. {\L}ojasiewicz,
``Sur la g$\acute{e}$om$\acute{e}$trie semi- et sous-analytique,''
{\it Ann. Inst. Fourier}, vol. 43, no. 5, pp. 1575-1595, 1993.

\bibitem{Tpcl}
T. P. Chen and L. L. Wang,
``Power-rate global stability of dynamical systems with unbounded time-varying delays,''
{\it IEEE Trans. Circuits Syst. II, Exp. Briefs}, vol. 54, pp. 705-709, Aug. 2007.

\bibitem{Tpc}
T. P. Chen and S. Amari,
``Stability of asymmetric Hopfield networks,''
{\it IEEE Trans. Neural Netw.}, vol. 12, pp. 159-163, Jan. 2001.

\bibitem{Wll}
W. L. Lu and J. Wang,
``Convergence analysis of a class of nonsmooth gradient systems,''
{\it IEEE Trans. Circuits Syst. I, Reg. Papers}, vol. 55, pp. 3514-3527, Dec. 2008.

\bibitem{Wllt}
W. L. Lu and T. P. Chen,
``New conditions on global stability of Cohen-Grossberg neural networks,''
{\it Neural Comput.}, vol. 15, no. 5, pp. 1173-1189, 2003.

\bibitem{Wzh1}
W. Zhu and Z. P. Jian,
``Event-Based Leader-following Consensus of Multi-agent Systems with Input Time Delay,''
{\it IEEE Trans. Autom. Control}, vol. 60, pp. 1362-1367, May. 2015.

\bibitem{Wzh2}
W. Zhu, Z. P. Jian, and G. Feng,
``Event-based consensus of multi-agent systems with general linear models,''
{\it Automatica}, vol. 50, no. 2, pp. 552-558, 2014.

\bibitem{Xwmd}
X. Wang and M. D. Lemmon,
``Self-triggered feedback control systems with finite-gain $\mathcal L_{2}$ stability,''
{\it IEEE Trans. Autom. Control}, vol. 45, pp. 452-467, Mar. 2009.

\bibitem{Xwm}
X. Wang and M. D. Lemmon,
``Event-triggering distributed networked control systems,''
{\it IEEE Trans. Autom. Control}, vol. 56, pp. 586-601, Mar. 2011.

\bibitem{Yfg}
Y. Fan, G. Feng, Y. Wang, and C. Song,
``Distributed event-triggered control of multi-agent systems with combinational measurements,''
{\it Automatica}, vol. 49, no. 2, pp. 671-675, 2013.

\bibitem{Zliu}
Z. Liu, Z. Chen, and Z. Yuan,
``Event-triggered average-consensus of multi-agent systems with weighted and direct topology,''
{\it J. Syst. Sci. Complex.}, vol. 25, no. 5, pp. 845-855, 2012.


\end{thebibliography}
\end{document}